\documentclass[runningheads]{llncs}

\usepackage{url}
\usepackage{epsfig}
\usepackage[hidelinks]{hyperref}
\usepackage{academicons}
\usepackage{graphicx}
\usepackage{amsmath}
\usepackage{amssymb}
\usepackage{subcaption}
\usepackage{microtype}
\usepackage[dvipsnames]{xcolor}

\DeclareMathOperator{\sigm}{sigm}

\DeclareMathOperator{\CrsEnt}{\mathcal{L}_{CE}}
\DeclareMathOperator{\Sep}{\mathcal{L}_{Sep}}
\DeclareMathOperator{\Clst}{\mathcal{L}_{Clst}}

\DeclareMathOperator{\xx}{\mathbf{x}}
\DeclareMathOperator{\zz}{\mathbf{z}}
\DeclareMathOperator{\pp}{\mathbf{p}}
\DeclareMathOperator{\ww}{\mathbf{w}}
\DeclareMathOperator{\hh}{\mathbf{h}}

\usepackage{wrapfig}
\usepackage{etoolbox}
\usepackage[misc]{ifsym}
\newcommand{\orcid}[1]{\href{https://orcid.org/#1}{\includegraphics[scale=0.025]{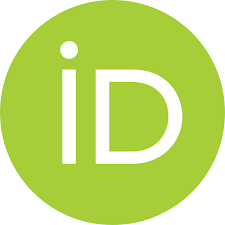}}}
\newcommand{\repthanks}[1]{\textsuperscript{\ref{#1}}}
\makeatletter
\patchcmd{\maketitle}
  {\def\thanks}
  {\let\repthanks\repthanksunskip\def\thanks}
  {}{}
\patchcmd{\@maketitle}
  {\def\thanks}
  {\let\repthanks\@gobble\def\thanks}
  {}{}
\newcommand\repthanksunskip[1]{\unskip{}}
\makeatother

\begin{document}

%%%%%%%%%%%%%%%%%%%%
\title{ProtoMIL: Multiple Instance Learning with Prototypical Parts for Whole-Slide Image Classification} %for Fine-Grained Interpretability}
\titlerunning{ProtoMIL: MIL with Prototypical Parts for WSI Classification}

% \author{Anonymous authors}
\toctitle{ProtoMIL: Multiple Instance Learning with Prototypical Parts for Whole-Slide Image Classification}
\tocauthor{Dawid~Rymarczyk}
\tocauthor{Adam~Pardyl}
\tocauthor{Jaros\l{}aw~Kraus}
\tocauthor{Aneta~Kaczy\'nska}
\tocauthor{Marek~Skomorowski}
\tocauthor{Bartosz~Zieli\'nski}
\author{Dawid Rymarczyk\inst{1,2}\orcidID{0000-0002-8543-5200}(\Letter)%\orcidID{0000-0002-8543-5200} 
\and 
Adam Pardyl\inst{1}\orcidID{0000-0002-3406-6732}\thanks{denotes equal contribution\protect\label{EQ}} \and
Jaros\l{}aw Kraus\inst{1}\orcidID{0000-0001-6904-1351}\repthanks{EQ} \and\\
Aneta Kaczy\'nska\inst{1}\orcidID{0000-0001-7571-8357}\repthanks{EQ} \and 
Marek Skomorowski \inst{1}\orcidID{0000-0002-1215-4379} \and
Bartosz Zieli\'nski \inst{1,2}\orcidID{0000-0002-3063-3621}}

% \authorrunning{Anonymous authors}
\authorrunning{D. Rymarczyk, et al.}% A. Kaczy\'nska, J. Kraus, A. Pardyl, M. Skomorowski, B. Zieli\'nski}
% {\tt\small bartosz.zielinski@uj.edu.pl}
% \and
% \institute{anonymous institute}
\institute{Faculty of Mathematics and Computer Science, Jagiellonian University,\\
6~\L{}ojasiewicza Street, 30-348 Krak\'ow, Poland
\and
Ardigen SA, 76~Podole Street, 30-394 Krak\'ow, Poland
\email{\{dawid.rymarczyk,adam.pardyl,jarek.kraus,\\aneta.kaczynska\}@student.uj.edu.pl, \{marek.skomorowski,bartosz.zielinski\}@uj.edu.pl} 
}

% \affil[e]{\tt\small denotes equal contribution
% }

\maketitle

%%%%%%%%%%%%%%%%%%%%
\begin{abstract}
The rapid development of histopathology scanners allowed the digital transformation of pathology. Current devices fastly and accurately digitize histology slides on many magnifications, resulting in whole slide images (WSI). However, direct application of supervised deep learning methods to WSI highest magnification is impossible due to hardware limitations. That is why WSI classification is usually analyzed using standard Multiple Instance Learning (MIL) approaches, that do not explain their predictions, which is crucial for medical applications.
In this work, we fill this gap by introducing ProtoMIL, a novel self-explainable MIL method inspired by the case-based reasoning process that operates on visual prototypes. Thanks to incorporating prototypical features into objects description, ProtoMIL unprecedentedly joins the model accuracy and fine-grained interpretability, as confirmed by the experiments conducted on five recognized whole-slide image datasets.

%Multiple Instance Learning (MIL) gains popularity in many real-life machine learning applications due to its weakly supervised nature. However, the corresponding effort on explaining MIL lags behind, and it is usually limited to presenting instances of a bag that are crucial for a particular prediction. In this paper, we fill this gap by introducing ProtoMIL, a novel self-explainable MIL method inspired by the case-based reasoning process that operates on visual prototypes. Thanks to incorporating prototypical features into objects description, ProtoMIL unprecedentedly joins the model accuracy and fine-grained interpretability, which we present with the experiments on five recognized MIL datasets.
\end{abstract}

% \todo{Uzupełnić keywordy}

%%%%%%%%%%%%%%%%%%%%
\section{Introduction}
\label{sec:intro}

A typical supervised learning scenario assumes that each data point has a separate label. However, in Whole Slide Image (WSI) classification, only one label is usually assigned to a gigapixel image due to the laborious and expensive labeling. Because of the hardware limitations, the direct application of supervised deep learning methods to WSI two highest magnification is impossible. That is why recent approaches~\cite{li2021dual} divide the WSI into smaller patches (instances) and process them separately to obtain their representations. Such representations form a bag of instances associated with only one label, and it is unspecified which instances are responsible for this label~\cite{foulds2010review}. This kind of problem, called Multiple Instance Learning (MIL)~\cite{dietterich1997solving}, appears in many medical problems, such as the diabetic retinopathy screening~\cite{quellec2012multiple,rani2016multiple}, bacteria clones identification using microscopy images~\cite{borowa2020classifying}, or identifying conformers responsible for molecule activity in drug design~\cite{straehle2014multiple,zhao2013drug}.

\begin{figure}[t]
  \centering
  \includegraphics[width=0.9\textwidth]{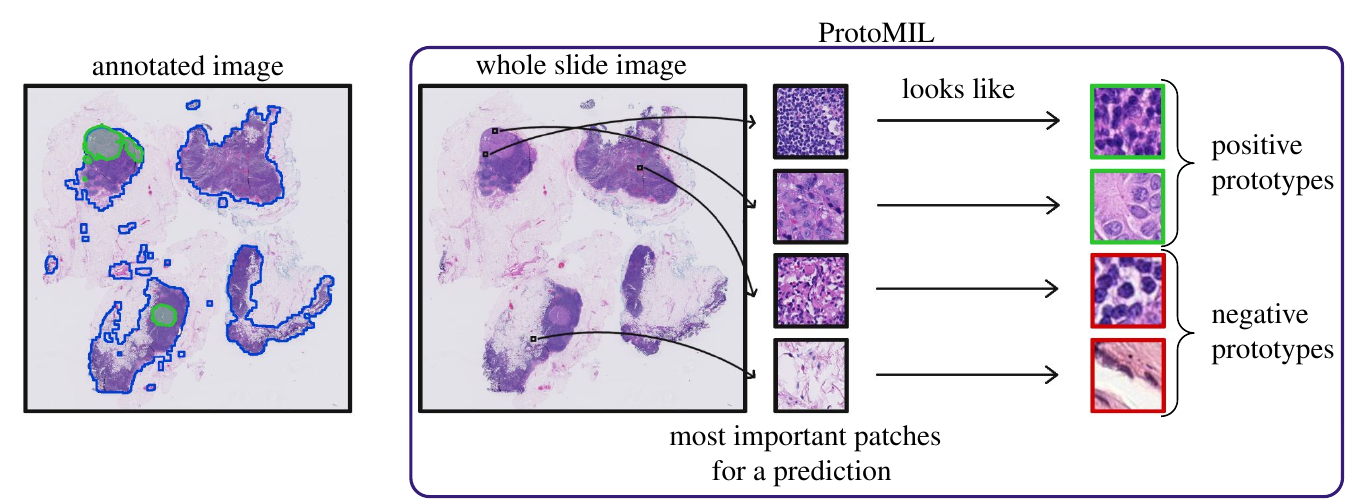}
  \caption{ProtoMIL divides the whole slide image into patches and analyzes their similarity to the reference prototypical parts that describe the given data class. As a result, it can provide a visual explanation of its prediction. One can observe that ProtoMIL identifies the most important patches with attention weights, that can appear both inside and outside a cancer region (marked as green and blue areas, respectively). Moreover, these patches are described by cancer or healthy tissue prototypes (corresponding to patches in green and red frames, respectively), showing their resemblance to the training examples.}
  \label{fig:overview}
\end{figure}

In recent years, with the rapid development of deep learning, MIL is combined with many neural network-based models~\cite{feng2017deep,ilse2018attention,li2021dual,lu2021data,rymarczyk2021kernel,shao2021transmil,shi2020loss,tu2019multiple,wang2018revisiting,yan2018deep}. Many of them embed all instances of the bag using a convolutional block of a deep network and then aggregate those embeddings. Moreover, some aggregation methods specify the most important instances that are presented to the user as prediction interpretation~\cite{ilse2018attention,li2021dual,lu2021data,rymarczyk2021kernel,shi2020loss}.
However, those methods usually only exhibit instances crucial for the prediction and do not indicate the cause of their importance. Naturally, there were attempts to further explain the MIL models~\cite{barnett2021iaia,borowa2020classifying,li2021multi}, but overall, they usually introduce additional bias into the explanation~\cite{rudin2019stop} or require additional input~\cite{li2021multi}.

To address the above shortcomings of MIL models, we introduce \textit{Prototypical Multiple Instance Learning} (ProtoMIL). It builds on case-based reasoning, a type of explanation naturally used by humans to describe their thinking process~\cite{kolodner2014case}. More precisely, we divide each WSI into patches and analyze how similar they are to a trainable set prototypical parts of positive and negative data classes, as defined in~\cite{chen2018looks}. Since, the prototypes are trainable, they are automatically derived by ProtoMIL. Then, we apply an attention pooling operator to accumulate those similarities over instances. As a result, we obtain bag-level representation classified with an additional neural layer.
This approach significantly differs from non-MIL approaches because it applies an aggregation layer and introduces a novel regularization technique that encourages the model to derive prototypes from the instances responsible for the positive label of a bag. The latter is a challenging problem because those instances are concealed and underrepresented. Lastly, the prototypical parts are pruned to characterize the data classes compactly. This results in detailed interpretation, where the most important patches according to attention weights are described using prototypes, as shown in Fig.~\ref{fig:overview}.

To show the effectiveness of our model, we conduct experiments on five WSI datasets: Bisque Breast Cancer~\cite{gelasca2008evaluation}, Colon Cancer~\cite{sirinukunwattana2016locality},  Camelyon16 Breast Cancer~\cite{camelyon16paper}, Lung cancer subtype identification TCGA-NSCLC~\cite{bakr2018radiogenomic} and Kidney cancer subtype classification~\cite{akin2016radiology}. Additionally, in the Supplementary Materials, we show the universal character of our model in different scenarios such as MNIST Bags \cite{ilse2018attention} and Retinopathy Screening (Messidor dataset)~\cite{decenciere2014feedback}. The results we obtain are usually on par with the current state-of-the-art models. However, at the same time, we strongly enhance interpretation capabilities with prototypical parts obtained from the training set. We made our code publicly available at \url{https://github.com/apardyl/ProtoMIL}.

The main contributions of this work are as follows:
\begin{itemize}
  \item Introducing the ProtoMIL method, which substantially improves the interpretability of existing MIL models by introducing case-based reasoning.
  \item Developing a training paradigm that encourages generating prototypical parts from the underrepresented instances responsible for the positive label of a bag.
\end{itemize}

The paper is organized as follows. In Section~\ref{sec:rw}, we present recent advancements in Multiple Instance Learning and deep interpretable models. In Section~\ref{sec:methods}, we define the MIL paradigms and introduce ProtoMIL. Finally, in Section~\ref{sec:results}, we present the results of conducted experiments, and Section~\ref{sec:conclusion} summarizes the work.

%%%%%%%%%%%%%%%%%%%%
\section{Related works}
\label{sec:rw}
% \todo{with good results make it more like whole-slide image}
Our work focuses on classification of whole slide images which is described using  Multiple Instance Learning (MIL) framework. Additionally, we develop an interpretable method which relates to eXplainable Artificial Intelligence (XAI). We briefly describe both fields in the following subsections.

\subsection{Multiple instance learning}

Before the deep learning era, models based on SVM, such as MI-SVM~\cite{andrews2002support}, were used for MIL problems. However, currently, MIL is addressed with numerous deep models. One of them, Deep MIML~\cite{feng2017deep}, introduces a sub-concept layer that is learned and then pooled to obtain a bag representation. Another example is mi-Net~\cite{wang2018revisiting}, which pools predictions from single instances to derive a bag-level prediction. Other architectures adapted to MIL scenarios includes capsule networks~\cite{yan2018deep}, transformers~\cite{shao2021transmil} and graph neural networks~\cite{tu2019multiple}. Moreover, many works focus on the attention-based pooling operators, like AbMILP introduced in~\cite{ilse2018attention} that weights the instances embeddings to obtain a bag embedding. This idea was also extended by combining it with mi-Net~\cite{li2021dual}, clustering similar instances~\cite{lu2021data}, self-attention mechanism~\cite{rymarczyk2021kernel}, and sharing classifier weights with pooling operator~\cite{shi2020loss}.
However, the above methods either do not contain an XAI component or only present the importance of the instances. Hence, our ProtoMIL is a step towards the explainability of the MIL methods.

\subsection{Explainable artificial intelligence}

There are two types of eXplainable Artificial Intelligence (XAI) approaches, post hoc and self-explaining methods~\cite{arya2019one}. Among many \textit{post hoc} techniques, one can distinguish saliency maps showing pixel importance~\cite{rebuffi2020there,selvaraju2017grad,selvaraju2019taking,simonyan2013deep} or concept activation vectors representing internal network state with human-friendly concepts~\cite{chen2020concept,ghorbani2019towards,kim2018interpretability,yeh2019completeness}. They are easy to use since they do not require any changes in the model architecture. However, their explanations may be unfaithful and fragile~\cite{adebayo2018sanity}.
Therefore \textit{self-explainable} models were introduced like Prototypical Part Network~\cite{chen2018looks} with a layer of prototypes representing the activation patterns. A similar approach for hierarchically organized prototypes is presented in~\cite{hase2019interpretable} to classify objects at every level of a predefined taxonomy. Moreover, some works concentrate on transforming prototypes from the latent space to data space~\cite{li2018deep} or focus on sharing prototypical parts between classes and finding semantic similarities~\cite{rymarczyk2020protopshare}. Other works~\cite{nauta2021neural} build a decision tree with prototypical parts in the nodes or learn disease representative features within a dynamic area~\cite{kim2021xprotonet}. Nonetheless, to our best knowledge, no fine-grained self-explainable method, like ProtoMIL, exists for MIL problems.

\begin{figure}[t!]
  \centering
%   \vspace{-3em}
  \includegraphics[width=0.59\textwidth]{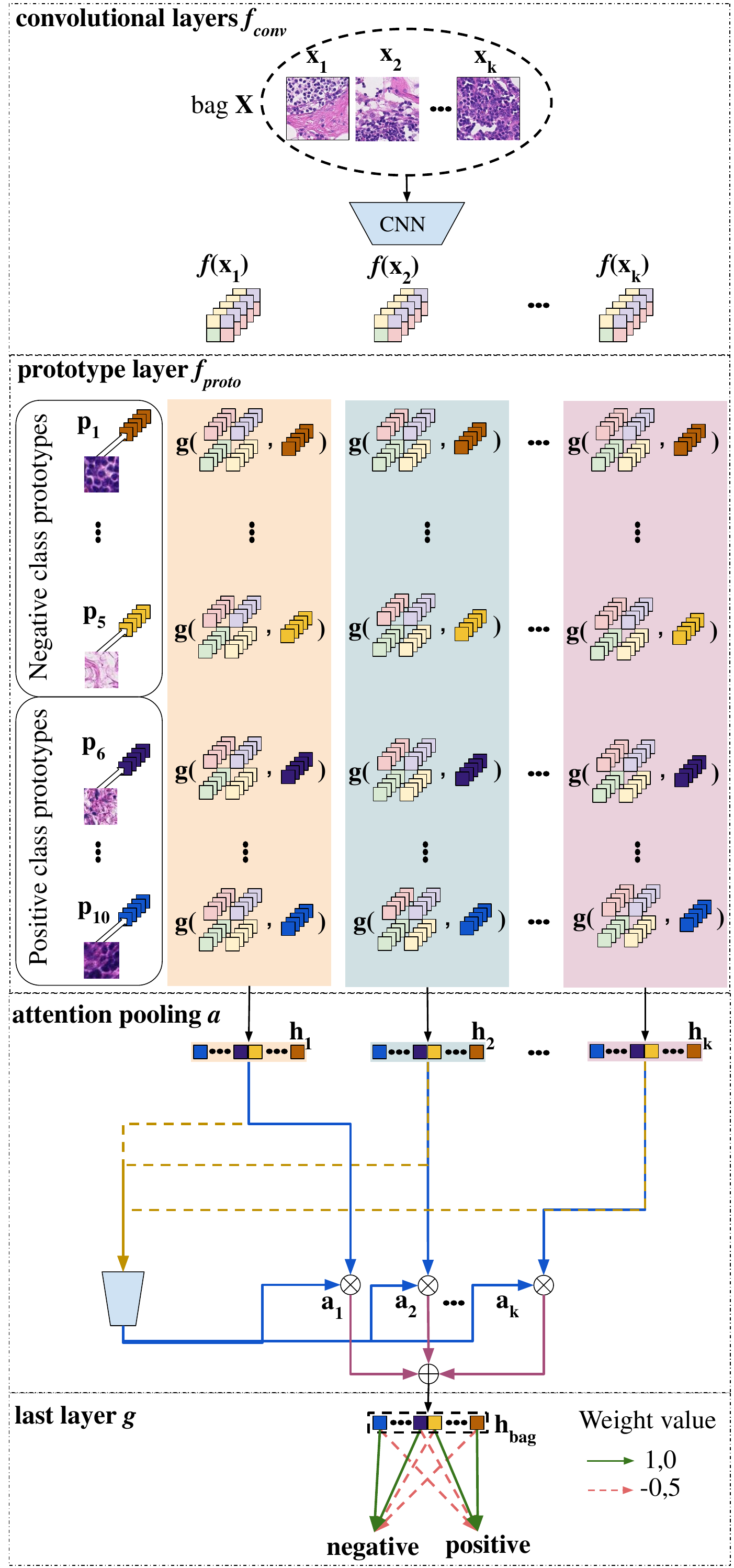}
  \caption{ProtoMIL passes a bag of patches through four modules. First, convolutional layer $f_{conv}$ generates embeddings for each patch. Then, the prototype layer $f_{proto}$ calculates similarities between patches representations and its prototypes. The similarities are aggregated using the attention pooling $a$ to obtain the bag similarity scores classified using the last layer $g$. Notice that particular colors in vectors $\mathbf{h_i}$ and $\mathbf{h_{bag}}$ correspond to prototypes similarities. %(brown and yellow for positive class, and dark and bright blue for the negative class).
  % \url{https://docs.google.com/drawings/d/1T9LknjyMtb9FV4Sk03c-G0pnpwQGCkpxEpw57aZDe2A/edit?usp=sharing}
  }
%   \vspace{-8em}
  
  \label{fig:protoMIL}
\end{figure}

%%%%%%%%%%%%%%%%%%%%
\section{ProtoMIL}
\label{sec:methods}

Due to the large resolution of whole slide images, which should not be scaled down due to loss of information, we first divide an image into patches. However, we do not know which patches correspond to the given disease state. Therefore, this problem boils down to Multiple Instance Learning (MIL), where there is a bag of instances (in our case patches) and only one label for the whole bag. This bag is passed trough the four modules of ProtoMIL (see Fig.~\ref{fig:protoMIL}): convolutional network $f_{conv}$, prototype layer $f_{proto}$, attention pooling $a$, and fully connected last layer $g$. Convolutional and prototype layers process single instances, whereas attention pooling and the last layer work on a bag level.
More precisely, given a bag of patches $X = \{\xx_1,\dots,\xx_k\}$, each $\xx \in X$ is forwarded through convolutional layers to obtain low-dimensional embeddings $F = \{f_{conv}(\xx_1),\dots,f_{conv}(\xx_k)\}$. As $f_{conv}(\xx) \in H\times W\times D$, for the clarity of description, let $Z_{\xx}=\{\zz_j\in f_{conv}(\xx) : \zz_j\in \mathbb{R}^D, j=1..HW\}$.
Then, the prototype layer computes vector $\hh$ of similarity scores~\cite{chen2018looks} between each embedding $f_{conv}(\xx)$ and all prototypes $\pp \in P$ as
\[
	\hh = \left(g(Z_{\xx}, \pp) = \max\limits_{\zz\in Z_{\xx}} \log\left(\tfrac{\|\zz - \pp\|^2 + 1}{\|\zz - \pp\|^2 + \varepsilon}\right)\right)_{\pp \in P}\; \text{ for }\; \varepsilon > 0.
\]
This results in a bag of similarity scores $H = \{\hh_1,\dots,\hh_k\}$, which we pass to the attention pooling~\cite{ilse2018attention} to obtain a single similarity scores for the entire bag
\begin{equation}
    \hh_{bag} = \sum_{i=1}^{k} a_i\hh_i,\; \text{ where }\;
    a_i = \frac
    { \exp \{ \ww^T (\tanh (\mathbf{V}\hh_i^T) \odot \sigm (\mathbf{U}\hh_i^T)\}}
    {\sum\limits_{j=1}^k \exp \{\ww^T (\tanh (\mathbf{V}\hh_j^T) \odot \sigm (\mathbf{U}\hh_j^T)\}},
    \label{eq:h_a}
\end{equation}
$\ww \in \mathbb{R}^{L \times 1}$,
$\mathbf{V} \in \mathbb{R}^{L \times M}$, and
$\mathbf{U} \in \mathbb{R}^{L \times M}$
are parameters, $\tanh$ is the hyperbolic tangent, $\sigm$ is the sigmoid non-linearity and $\odot$ is an element-wise multiplication. Note that weights $a_i$ sum up to 1, and thus the formula is invariant to the size of the bag.
Such representation is then sent to the last layer to obtain the predicted label $\check{y} = g(h_{bag})$ as in~\cite{chen2018looks}.

\paragraph{Regularization.}
In MIL, the instances responsible for the positive label of a bag are underrepresented. Hence, training ProtoMIL without additional regularizations can result in a prototype layer with only prototypes of a negative class. That is why we introduce a novel regularization technique that encourages the model to derive positive prototypes.
For this purpose, we introduce the loss function composed of three components
\[
\CrsEnt(\check{y}, y) + 
\lambda_1 \Clst + \lambda_2 \Sep,
\]
where $\check{y}$ and $y$ denotes respectively the predicted and ground truth label of bag $X$, $\CrsEnt$ corresponds to cross-entropy loss, while
\[
\Clst = \frac{1}{|X|} \sum_{\xx_i \in X} a_i \min_{\pp \in P^y} \min_{\zz \in Z_{\xx_i}} \lVert{\zz-\pp}\rVert_2^2,
\]
\[
\Sep = - \frac{1}{|X|} \sum_{\xx_i \in X} a_i \min_{\pp \notin P^y} \min_{\zz \in Z_{\xx_i}} \lVert{\zz-\pp}\rVert_2^2,
\]
where $P^y$ is a set of prototypes assigned to class $y$.
Comparing to~\cite{chen2018looks}, components $\Clst$ and $\Sep$ additionally use $a_i$ from Equation~\ref{eq:h_a}. As a result, we encourage the model to create more prototypes corresponding to positive instances, which usually have higher $a_i$ values.

%%%%%%%%%%%%%%%%%%%%

\section{Experiments}
\label{sec:results}

We test our ProtoMIL approach on five datasets, for which we train the model from scratch in three steps: (i) \textit{warmup} phase with training all layers except the last one, (ii) prototype projection, (iii) and fine-tuning with fixed $f_{conv}$ and $f_{proto}$. Phases (ii) and (iii) are repeated several times to find the most optimal set of prototypes.
All trainings use Adam optimizer for all layers with $\beta_1 = 0.99$, $\beta_2 = 0.999$, weight decay $0.001$, and batch size $1$. Additionally, we use an exponential learning rate scheduler for the \textit{warmup} phase and a step scheduler for prototype training. All results are reported as an average of all runs with a standard error of the mean. In the subsequent subsections, we describe experiment details and results for each dataset.

Across all datasets we use convolutional block from ResNet-18 followed by two additional $1 \times 1$ convolutions as the convolutional layer $f_{conv}$. We use ReLU as the activation function for all convolutional layers except the last layer, for which we use the sigmoid activation function. The prototype layer stores prototypes shared across all bags, while the attention layer implements AbMILP. The last layer is used to classify the entire bag. Weights between similarity scores of prototypes corresponding class logit are initialized with $1$, while other connections are set to $-0.5$ as in~\cite{chen2018looks}. Together with the specific training procedure, such initialization results in a positive reasoning process (we rather say ``this looks like that'' instead of saying ``this does not look like that'').

\subsection{Bisque Breast Cancer and Colon Cancer datasets}
\paragraph{Experiment details.}

We experiment on two histological datasets: Colon Cancer and Bisque Breast Cancer. The former contains $100$ H\&E images with $22,444$ manually annotated nuclei of four different types: epithelial, inflammatory, fibroblast, and miscellaneous. To create bags of instances, we extract $27\times27$ nucleus-centered patches from each image, and the goal is to detect if the bag contains one or more epithelial cells, as colon cancer originates from them. On the other hand, the Bisque dataset consists of $58$ H\&E breast histology images of size $896\times768$, out of which $32$ are benign, and $26$ are malignant (contain at least one cancer cell). Each image is divided into $32\times32$ patches, resulting in $672$ patches per image. Patches with at least $75\%$ of the white pixels are discarded, resulting in $58$ bags of various sizes.

We apply extensive data augmentation for both datasets, including random rotations, horizontal and vertical flipping, random staining augmentation, staining normalization, and instance normalization. We use ResNet-18 convolutional parts with the first layer modified to $3\times3$ convolution with stride $1$ to match the size of smaller instances. We set the number of prototypes per class to $10$ with a size of $128\times2\times2$. Warmup, fine-tuning, and end-to-end training take $60$, $20$, and $20$ epochs, respectively. $10$-fold cross-validation with $1$ validation fold and $1$ test fold is repeated $5$ times.

\paragraph{Results.}
Table \ref{table:bisque_colon} presents our results compared to both traditional and attention-based MIL models. On the Bisque dataset, our model significantly outperforms all baseline models. However, due to the small size of the Colon Cancer dataset, ProtoMIL overfits, resulting in poorer AUC than attention-based models. Nevertheless, in both cases, ProtoMIL provides finer explanations than all baseline models (see Fig.~\ref{fig:mnist_small_matrix} and Supplementary Materials).

\begin{table}[t]
\caption{Results for small histological datasets, where ProtoMIL significantly outperforms baseline methods on the Bisque dataset. However, it achieves worse results for the Colon Cancer dataset, probably due to its small size. Additionally, interpretability of the methods is noted and further discussed in Section~\ref{sec:int}. Notice that values for comparison indicated with ``*'' and ``**'' comes from~\cite{ilse2018attention} and~\cite{rymarczyk2021kernel}, respectively.}
\centering
\scriptsize % \scriptsize or \footnotesize or \small
\begin{tabular}{|c||c|c||c|c||c|}
\hline
& \multicolumn{2}{c||}{\textsc{Bisque}} & \multicolumn{2}{c|}{\textsc{Colon Cancer}} & \\ [0.5ex]
\cline{2-5}
\textsc{Method} & \textsc{Accuracy} & \textsc{AUC} & \textsc{Accuracy} & \textsc{AUC} & \textsc{Inter.} \\ [0.5ex]
\hline
\textsc{instance+max*} & $61.4\%\pm 2.0\%$ & $0.612\pm 0.026$ & $84.2\%\pm 2.1\%$ & $0.914\pm 0.010$ & + \\ [1ex]
\textsc{instance+mean*} & $67.2\%\pm 2.6\%$ & $0.719\pm 0.019$ & $77.2\%\pm 1.2\%$ & $0.866\pm 0.008$ & - \\ [1ex]
\textsc{embedding+max*} & $60.7\%\pm 1.5\%$ & $0.650\pm 0.013$ & $82.4\%\pm 1.5\%$ & $0.918\pm 0.010$ & - \\ [1ex]
\textsc{embedding+mean*} & $74.1\%\pm 2.3\%$ & $0.796\pm 0.012$ & $86.0\%\pm 1.4\%$ & $0.940\pm 0.010$ & -\\ [1ex]
\textsc{AbMILP*} & $71.7\%\pm 2.7\%$ & $0.856\pm 0.022$ & $88.4\%\pm 1.4\%$ & $0.973\pm 0.007$ & ++ \\ [1ex]
\textsc{SA-AbMILP**} & $75.1\%\pm 2.4\%$ & $0.862\pm 0.022$ & \textbf{90.8\%} $\pm$ \textbf{1.3\%} & \textbf{0.981} $\pm$ \textbf{0.007} & + \\ [1ex]
\hline
\textsc{ProtoMIL (our)} & \textbf{76.7\%} $\pm$ \textbf{2.2\%} & \textbf{0.886} $\pm$ \textbf{0.033} & $81.3\%\pm 1.9\%$ & $0.932\pm 0.014$ & +++\\ [1ex]
\hline
\end{tabular}
\label{table:bisque_colon}
\end{table}

\begin{figure}[ht]
  \centering
  \includegraphics[width=0.9\textwidth]{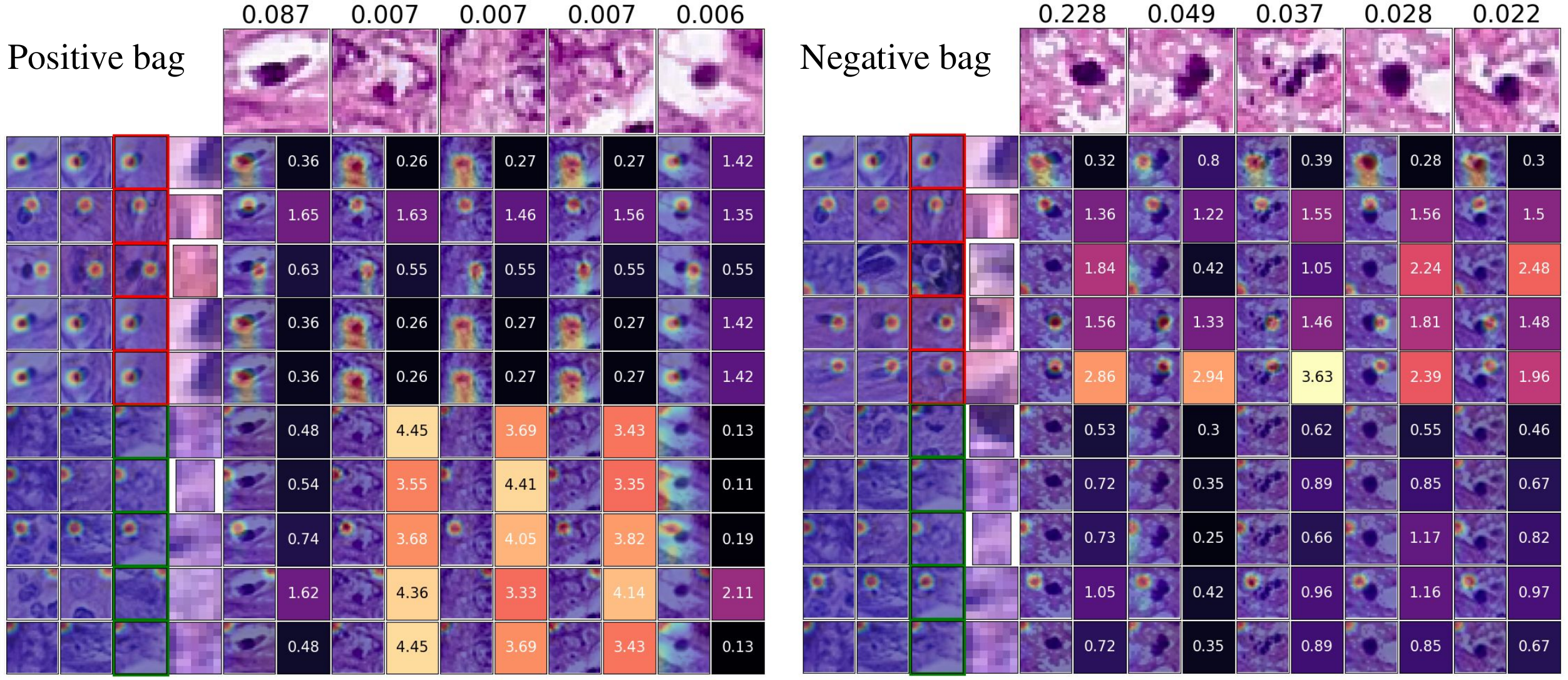}
  \caption{Similarity scores between five crucial instances of a bag (columns) and ten prototypical parts (rows) for a positive and negative bag (left and right side, respectively) from the Colon Cancer bags. Each prototypical part is represented by a part of the training image and three nearest training patches, and each instance is represented by the patch and the value of its attention weight $a_i$. Moreover, each cell contains a similarity score and a heatmap corresponding to prototype activation. One can observe that instances of a negative bag usually activate prototypes of a negative class (four upper prototypes in red brackets), while the instances of positive bags mostly activate positive prototypes (four bottom prototypes in green brackets).}
  \label{fig:mnist_small_matrix}
\end{figure}

\begin{table}[t]
\caption{Our ProtoMIL achieves state-of-the-art results on the Camelyon16 dataset in terms of AUC metric, surpassing even the transformer-based architecture. Moreover, it is competitive on TCGA-NSCLC and slightly worse on TCGA-RCC, with a small drop of accuracy and AUC compared to TransMIL. Additionally, interpretability of the methods is noted and further discussed in Section~\ref{sec:int}. Notice that values for comparison marked with ``*'' and ``**'' are taken from~\cite{li2021dual} and~\cite{shao2021transmil}, respectively.}
\centering
\scriptsize % \scriptsize or \footnotesize or \small
\begin{tabular}{|c||c|c||c|c||c|c||c|}
\hline
& \multicolumn{2}{c||}{\textsc{Camelyon16}} & \multicolumn{2}{c||}{\textsc{TCGA-NSCLC}} & \multicolumn{2}{c|}{\textsc{TCGA-RCC}} & \\ [0.5ex]
\cline{2-7}
\textsc{Method} & \textsc{Accuracy} & \textsc{AUC} & \textsc{Accuracy} & \textsc{AUC} & \textsc{Accuracy} & \textsc{AUC} & \textsc{Inter.}\\ [0.5ex]
\hline
\textsc{instance+mean*} & 79.84\% & 0.762 & 72.82\% & 0.840 &90.54\% & 0.978 & -\\ [1ex]
\textsc{instance+max*} & 82.95\% & 0.864 & 85.93\% & 0.946 & 93.78\% & 0.988  & +\\ [1ex]
\textsc{MILRNN*} & 80.62\% & 0.807 & 86.19\% & 0.910 & - & -  & -\\ [1ex]
\textsc{ABMILP*} & 84.50\% & 0.865 & 77.19\% & 0.865 & 89.34\% & 0.970  & ++\\ [1ex]
\textsc{DSMIL*} & 86.82\% & 0.894 & 80.58\% & 0.892 & 92.94\% & 0.984  & ++\\ [1ex]
\textsc{CLAM-SB**} & 87.60\% & 0.881 & 81.80\% & 0.881 & 88.16\% & 0.972  & +\\ [1ex]
\textsc{CLAM-MB**} & 83.72\% & 0.868 & 84.22\% & 0.937 & 89.66\% &  0.980  & +\\ [1ex]
\textsc{TransMIL**} & \textbf{88.37\%} & 0.931 & \textbf{88.35\%} & \textbf{0.960} & \textbf{94.66}\% &  \textbf{0.988}  & +\\ [1ex]
\hline
\textsc{ProtoMIL} (our) & 87.29\% & \textbf{0.935} & 83.66\% & 0.918 & 92.79\% & 0.961  & +++\\ [1ex]
\hline
\end{tabular}
\label{table:camelyon_results}
\end{table}

\begin{figure}[h]
  \centering
  \includegraphics[width=0.49\textwidth]{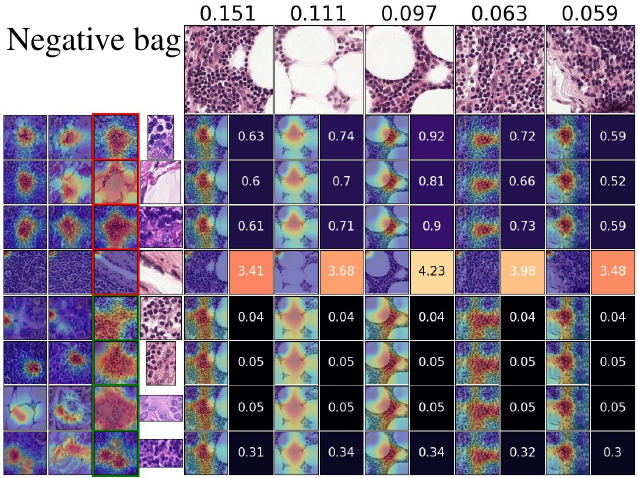}
  \caption{ Similarity scores between five crucial instances of a bag (columns) and eight prototypical parts (rows) for a negative bag from the Camelyon16 dataset. One can observe that ProtoMIL strongly activates only one prototype and focuses mainly on nuclei when analyzing the healthy parts of the tissue. Please refer to Fig.~\ref{fig:mnist_small_matrix} for a detailed description of the visualization.
  %Example of interpretation matrix for Camelyo dataset for negative class example. One can observe that only a small subset of prototypes is activated. We can see clear separation between activation of positive class prototypes and negative one. Also, we can see when the model takes into consideration nuclei and when it basis its decision on healthy parts of the tissue.
  }
  \label{fig:camelyon_neg_matrix_small}
\end{figure}

\begin{figure}[ht]
  \centering
  \includegraphics[width=0.49\textwidth]{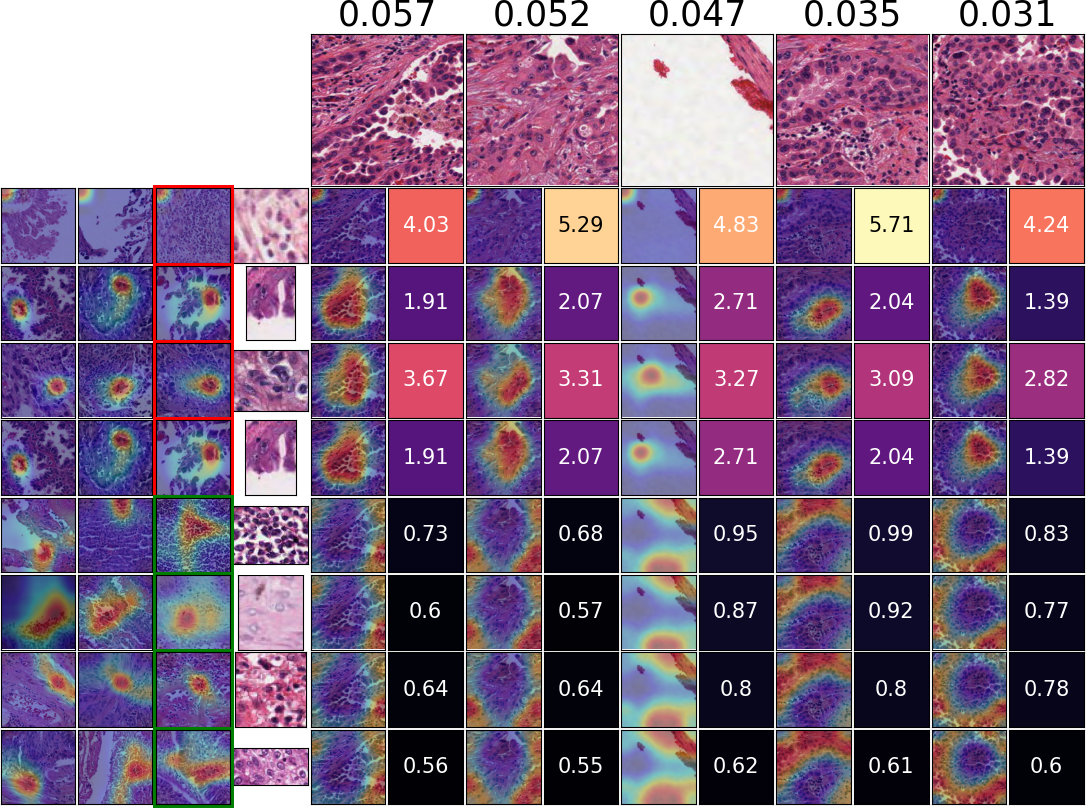}
  \caption{Similarity scores between five crucial instances of a bag (columns) and eight prototypical parts (rows) for a LUAD type bag from the TCGA-NSCLC dataset.
  }
  \label{fig:nsclc_matrix_small}
\end{figure}

\subsection{Camelyon16 dataset}

\paragraph{Experiment details.}
The Camelyon16 dataset~\cite{camelyon16paper} consists of $399$ whole-slide images of breast cancer samples, each labeled as \textit{normal} or \textit{tumor}. We create MIL bags by dividing each slide $20x$ resolution image into $224\times224$ patches, rejecting patches that contain more than $70\%$ of background. This results in $399$ bags with a mean of $8,871$ patches and a standard deviation of $6,175$. Moreover, $20$ largest bags are truncated to $20,000$ random patches to fit into the memory of a GPU. The positive patches are again highly imbalanced, as only less than $10\%$ of patches contain tumor tissue.

Due to the size of the dataset, we preprocess all samples using a ResNet-18 without two last layers, pre-trained on various histopathological images using self-supervised learning from~\cite{ciga2020self}. The resulting embeddings are fed into our model to replace the feature backbone net. ProtoMIL is trained for $50$, $40$, and $10$ epochs in warmup, fine-tuning, and end-to-end training, respectively. The number of prototypes per class is limited to $5$ with no data augmentation. The experiments are repeated $5$ times with the original train-test split.

\paragraph{Results.}
We compare ProtoMIL to other state-of-the-art MIL techniques, including both traditional mean and max MIL pooling, RNN, attention-based MIL pooling, and transformer-based MIL pooling~\cite{shao2021transmil}. ProtoMIL performs on par in terms of accuracy and slightly outperforms other models on AUC metric (Table~\ref{table:camelyon_results}) while providing a better understanding of its decision process, as presented in Fig.~\ref{fig:camelyon_neg_matrix_small} and Supplementary Materials.

\subsection{TCGA-NSCLC dataset}

\paragraph{Experiment details.}
TCGA-NSCLC includes two subtype projects, i.e., Lung Squamous Cell Carcinoma (TGCA-LUSC) and Lung Adenocarcinoma (TCGA-LUAD), for a total of 956 diagnostic WSIs, including 504 LUAD slides from 478 cases and 512 LUSC slides from 478 cases. We create MIL bags using WSI Segmentation and Patching from~\cite{lu2021data} with default parameters, except patch-level parameter set to 1. Each slide image is cropped into a series of 224×224 patches. This results in $1,016$ bags with a mean of $3,961$ patches. 
We randomly split the data in the ratio of train:valid:test equal 60:15:25 and assure that there is no case overlap between the sets, and use the same ProtoMIL settings as in the Camelyon16 dataset are used. The results are reported for 4-fold cross-validation.

\paragraph{Results.}
Results for the TCGA-NSCLC dataset are presented in Table~\ref{table:camelyon_results} alongside results of other state-of-the-art approaches from~\cite{shao2021transmil}. ProtoMIL performs slightly lower on the Area Under the ROC Curve (AUC) and accuracy metrics than the powerful transformer-based model TransMIL but still is competitive to other CNN-based approaches. However, the advantage of ProtoMIL is its capability to provide a detailed explanation of predictions as presented in Fig.~\ref{fig:nsclc_matrix_small} and Supplementary Materials.

\subsection{TCGA-RCC dataset}

\paragraph{Experiment details.}
TCGA-RCC consists of three unbalanced classes: Kidney Chromophobe Renal Cell Carcinoma (TGCA-KICH, 111 slides from 99 cases), Kidney Renal Clear Cell Carcinoma (TCGA-KIRC, 489 slides from 483 cases), and Kidney Renal Papillary Cell Carcinoma (TCGA-KIRP, 284 slides from 264 cases) for a total of 884 WSIs.
We create MIL bags using WSI Segmentation and Paching from~\cite{lu2021data} with default parameters and a patch-level parameter set to 1. Each slide image is cropped into a series of 224×224 patches. This results in 884 bags with a mean of $4,309$ patches. A separate model is trained for each class, and scores are averaged for all classes.
Other experiment settings are identical as for TCGA-NSCLC described above.

\paragraph{Results.}
We compare ProtoMIL to other state-of-the-art MIL techniques, including both traditional mean and max MIL pooling, attention-based MIL pooling, and transformer-based MIL pooling~\cite{shao2021transmil}. ProtoMIL performs on par in terms of accuracy and AUC metric (Table~\ref{table:camelyon_results}) while providing a better understanding of its decision process, as presented in Supplementary Materials.

\begin{table*}[t!]
\caption{
The influence of ProtoMIL pruning on the accuracy and AUC score. One can notice that even though the pruning removes around $30\%$ of the prototypes, it usually does not noticeably decrease the AUC and accuracy of the model.}
\centering
\scriptsize % \scriptsize or \footnotesize or \small
\begin{tabular}{|c||c|c|c||c|c|c|}
\hline
& \multicolumn{3}{c||}{\textsc{Before pruning}} & \multicolumn{3}{c|}{\textsc{After pruning}} \\ [0.5ex]
\cline{2-7}
\textsc{Dataset} & \textsc{Proto. \#} & \textsc{Accuracy} & \textsc{AUC} & \textsc{Proto. \#} & \textsc{Accuracy} & \textsc{AUC} \\ [0.5ex]
\hline
% \textsc{MNIST Bags 500} & 20 $\pm$ 0 & 99.2\% $\pm$ 0.1\% & 0.999 $\pm$ 0.001 & 14.12 $\pm$ 0.28 & 99.2\% $\pm$ 0.1\% & 0.999 $\pm$ 0.001 \\ [1ex]
\textsc{Bisque} & 20 $\pm$ 0 & 76.7\% $\pm$ 2.2\% & 0.886 $\pm$ 0.033 & 13.6 $\pm$ 0.25 & 73.0\% $\pm$ 2.4\% & 0.867 $\pm$ 0.022 \\ [1ex]
\textsc{Colon Cancer} & 20 $\pm$ 0 & 81.3\% $\pm$ 1.9\% & 0.932 $\pm$ 0.014 & 15.69 $\pm$ 0.34 & 81.8\% $\pm$ 2.4\% & 0.880 $\pm$ 0.022\\ [1ex]
% \textsc{Messidor} & 20 $\pm$ 0 & 70.0\% $\pm$ 0.9\% & 0.692 $\pm$ 0.012 & 16.70 $\pm$ 1.86 & 64.7\% $\pm$ 1.3\% & 0.717 $\pm$ 0.013 \\ [1ex]
\textsc{Camelyon16} & 10 $\pm$ 0 & 87.3\% $\pm$ 1.2 \% & 0.935 $\pm$ 0.007 & 6.4 $\pm$ 0.24 & 85.9\% $\pm$ 1.5\% & 0.937 $\pm$ 0.007 \\ [1ex]
\textsc{TCGA-NSCLC} & 10 $\pm$ 0 & 83.66\% $\pm$  1.6\% & 0.918 $\pm$ 0.003 & 7.6 $\pm$ 1.2 & 81.1\% $\pm$ 1.4\% & 0.880 $\pm$ 0.003  \\ [1ex]
\textsc{TCGA-RCC} & 10 $\pm$ 0 & 94.66\% $\pm$  1.0\% & 0.988 $\pm$ 0.009 & 6.2 $\pm$ 1.2 & 91.5\% $\pm$ 1.2\% & 0.955 $\pm$ 0.006  \\ [1ex]
\hline
\end{tabular}

\label{table:pruning}
\end{table*}
\subsection{Pruning}

\paragraph{Experiment details.}
We run prototype pruning experiments on all the datasets to remove not class-specific prototypical parts and check their influence on the model performance. For each of them, we use the model trained in the previously described experiments. As pruning parameters, we use $k=6$ and $l=40\%$ and fine-tuned for $20$ epochs. Details about pruning operation are described in the Supplementary Materials.

\paragraph{Results.}
The accuracy and AUC in respect to the number of prototypes before and after pruning are presented in Table \ref{table:pruning}. 
For all datasets, the number of prototypes after pruning has decreased around $30\%$ on average. However, it does not result in a noticeable decrease in accuracy or AUC, except for Colon Cancer, where we observe a significant drop in AUC. Most probably, it is caused by the high visual resemblance of nuclei patches (especially between \textit{epithelial} and \textit{miscellaneous}) that after prototype projection may be very close to each other in the latent space. 

\subsection{Interpretability of MIL methods}\label{sec:int}

Column \textit{Inter.} in Tables~\ref{table:bisque_colon}, and~\ref{table:camelyon_results} indicates how interpretable are the considered models. Instances and embeddings-based methods, except instance-max, are not interpretable, similarly to MILRNN, since they lose information about instances crucial for the prediction. On the other hand, the AbMILP~\cite{ilse2018attention} identifies crucial instances within a bag and can present the local explanation to the users. However, other attention-based methods, such as SA-AbMILP~\cite{rymarczyk2021kernel}, TransMIL~\cite{shao2021transmil} and CLAMs~\cite{lu2021data} perform additional operations, like self-attention, requiring more effort from the user to analyze the explanation. That is why those methods have been assigned with lower interpretability. Moreover, DS-MIL~\cite{li2021dual}  finds a decision boundary on the bag level and can produce a more detailed explanation than AbMILP, but only for a single prediction (local explanations). In contrast, the ProtoMIL can produce both local (see Figure~\ref{fig:mnist_small_matrix}) and global explanations (see Supplementary Materials).

%%%%%%%%%%%%%%%%%%%%
\section{Discussion and conclusions}\label{sec:conclusion}

In this work, we introduce Prototypical Multiple Instance Learning (ProtoMIL), a method for Whole Slide Image classification that incorporates a case-based reasoning process into the attention-based MIL setup. In contrast to existing MIL methods, ProtoMIL provides a fine-grained interpretation of its predictions. For this purpose, it uses a trainable set of prototypical parts correlated with data classes.
The experiments on five datasets confirm that introducing fine-grained interpretability does not reduce the model's effectiveness, which is still on par with the current state-of-the-art methodology. Moreover, the results can be presented to the user with a novel visualization technique. 

The experiments show that ProtoMIL can be applied to a challenging problem like Whole-Slide Image classification. Therefore, in future works, we plan to generalize our method to multi-label scenarios and multimodal classification problems since WSI often comes with other medical data like CT and MRI.

\subsection{Limitations}
ProtoMIL limitations are inherited from the other prototype-based models, such as non-obvious prototype meaning. Ergo, prototype projection might still result in uncertainty on which attributes it represents. However, there are methods mitigating these, e.g. explainer defined in~\cite{nauta2021looks}. 

\subsection{Negative impact}
Our solution is based on prototypical parts that are susceptible to different types of adversarial attacks such as~\cite{hoffmann2021looks}. That is why practitioners shall address this risk in a deployed system with ProtoMIL. What is more, it may be used in information war to disinform societies when prototypes are obtained with spoiled data or are shown without appropriate comment, especially in fields like medicine.

\section{Acknowledgments}

This work was founded by the POIR.04.04.00-00-14DE/18-00 project carried out within the Team-Net programme of the Foundation for Polish Science cofinanced by the European Union under the European Regional Development Fund. This research was funded by the National Science Centre, Poland (research grant no. 2021/41/B/ST6/01370). For the purpose of Open Access, the authors have applied a CC-BY public copyright licence to any Author Accepted Manuscript (AAM) version arising from this submission.

{
\bibliographystyle{splncs04}
\bibliography{egbib}
}

\section*{Supplementary Materials}

In this Supplementary Materials, we present additional details on the ProtoMIL model and similarity scores visualizations with more instances and prototypes for all datasets considered in our experiments.

\section{ProtoMIL}
\subsection{Prototypes projection.}
Prototypes projection is an important step in the training procedure because it visualizes the prototypes using training patches. For this purpose, it replaces every learned prototype with the nearest training patch from the bag with the same label as the prototype class. The prototype $\mathbf{p^c}$ of class $c$ (negative or positive) can be replaced using the following formula
\[
\begin{gathered}
\mathbf{p^c} \leftarrow \arg\min_{\mathbf{z} \in Z} \lVert{\mathbf{z}-\mathbf{p^c}}\rVert_2,
\end{gathered}
\]
where $Z = \{\mathbf{z} \in Z_\mathbf{x} | \mathbf{x} \in X \land y = c\}$ and $y$ is a label of bag $X$.

\subsection{Pruning.}
During the prototype projection, every prototype is replaced with the representation of the nearest training patch from the bag with the same label. Generally, the representations of the nearest training patches correspond to the same label. However, in some cases, the nearest patches of a prototype correspond to more than one class. It is especially problematic in highly unbalanced datasets, frequently occurring in MIL tasks. To remove such misleading prototypes, we extend the prototype pruning algorithm from~\cite{chen2018looks} to work in the MIL scenario. More precisely, we find $k$-nearest training patches for each prototype $p_i^c$ belonging to class $c$. If out of those $k$ patches less than $r$ belong to bags labeled with class $c$, we assume that this prototype is not determinant and remove it. Moreover, in contrast to~\cite{chen2018looks}, we automatically select $r$ to remove up to $l\%$ of prototypes ($l$ and $k$ are selected so that both classes still contain prototypes, and the drop in training accuracy is minimal). Finally, we fine-tuned attention and the final layers to compensate for the prototype removal.
% {\small
% \bibliographystyle{ieee_fullname}
% \bibliography{egbib}
% }

\section{Additional results}

\subsection{MNIST Bags}
\paragraph{Experiment details.}
We experiment with the MNIST dataset, for which we generate the bags like proposed in~\cite{ilse2018attention}. Namely, a single bag contains grayscale images randomly sampled from the MNIST dataset. The bags' sizes are chosen using a normal distribution with a mean of $100$ and a standard deviation of $20$. A bag is considered positive if it contains at least one image labeled as ``$9$''. There are equal numbers of positive and negative bags.
Notice that even though such dataset is class-balanced, it contains only $5\%$ of images labeled as ``$9$'' ($10\%$ instances in the positive bags).
We test ProtoMIL for different size of dataset (50, 100, 200, 300, 400, 500 bags). Every experiment is run with random 10-fold cross-validation and repeated five times with a different seed to obtain mean AUC as the evaluation metric.
We train a model for $30$, $20$, and $10$ epochs for warmup, fine-tuning, and end-to-end training, respectively. The number of prototypes per class is set to 10, with prototype size $64\times2\times2$ (determined experimentally).

\begin{figure}[t]
  \centering
  \includegraphics[width=0.49\textwidth]{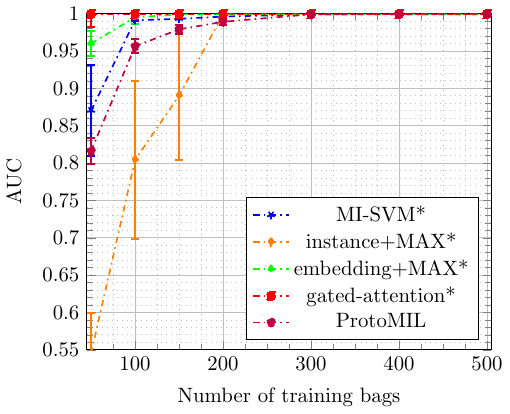}
  \caption{Results for ProtoMIL and baseline MIL approaches on the MNIST Bags dataset depending on the number of training bags (x axis) using the AUC metric (y axis). One can observe that ProtoMIL achieves state-of-the-art results with a larger number of samples.}
  \label{fig:mnist_auc}
\end{figure}

\paragraph{Results.}

\begin{figure}
  \centering
  \includegraphics[width=0.33\textwidth]{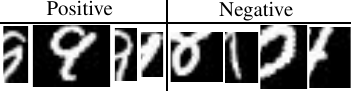}
  \caption{Sample positive and negative prototypes of ProtoMIL trained on the MNIST Bags dataset. Notice that the positive prototypes correspond to parts of ``$9$'' while the negative prototypes contain parts of the other digits (like ``$8$'' or ``$4$''). It is expected because a bag is considered positive if it contains at least one image of ``$9$''.}
  \label{fig:mnist_prototypes}
\end{figure}

We compare our model to baseline MIL pooling methods from~\cite{ilse2018attention}. As shown, our ProtoMIL approach requires slightly more samples to achieve AUC scores competitive to the regular models (Figure \ref{fig:mnist_auc}). However, as presented in Figure \ref{fig:mnist_small_matrix}a, it increases model interpretability by finding distinct parts of images and match them with intuitive positive and negative prototypes (see Figure \ref{fig:mnist_prototypes}).

\begin{figure*}
  \centering
  \includegraphics[width=0.8\textwidth]{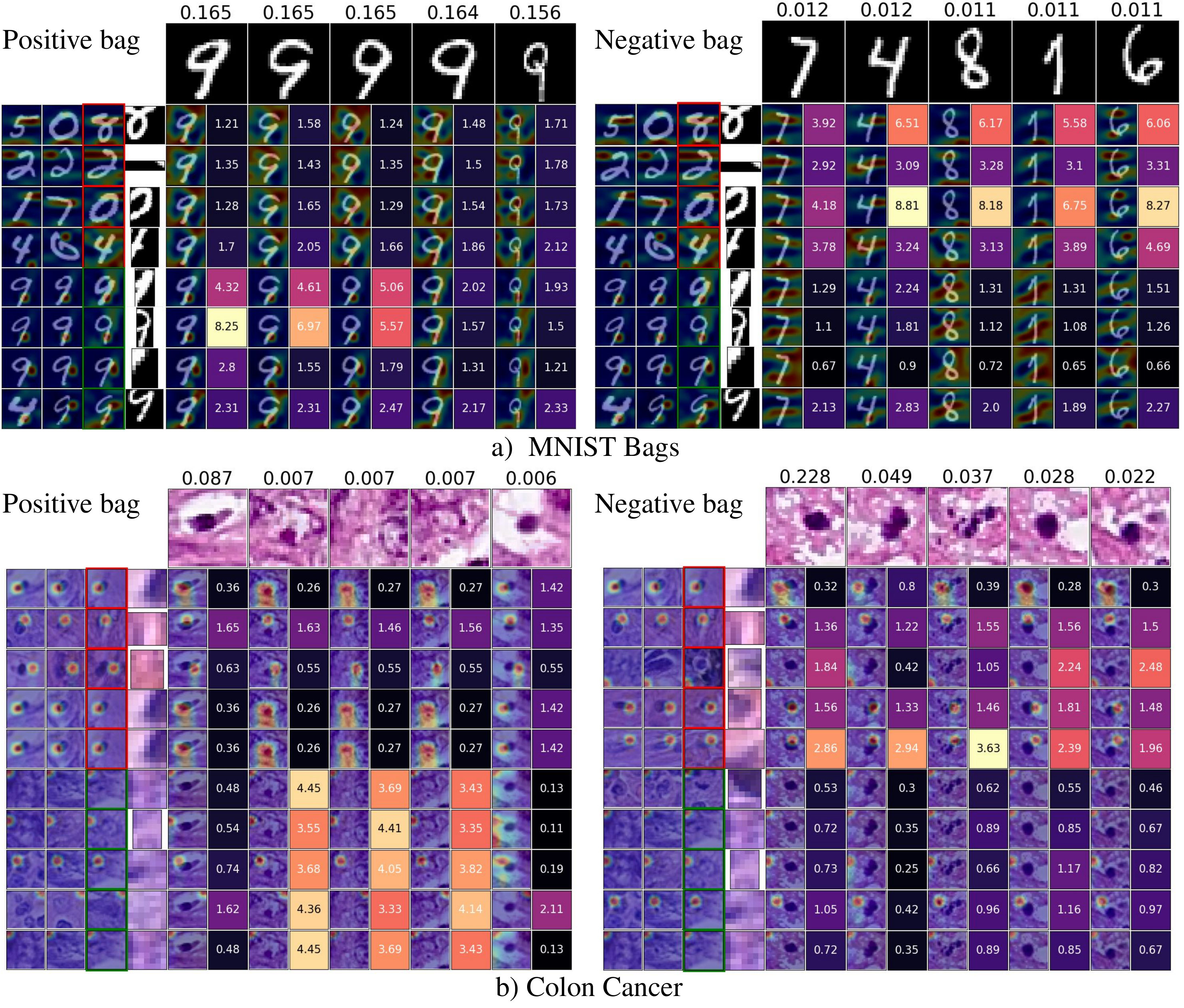}
  \caption{Similarity scores between five crucial instances of a bag (columns) and eight or ten prototypical parts (rows) for a positive and negative bag (left and right side, respectively) from the MNIST Bags (a) and Colon Cancer datasets (b). Each prototypical part is represented by a part of image and three nearest training patches, and each instance is represented by the image and the value of attention weight $a_i$. Moreover, each cell contains a similarity score and a heatmap corresponding to prototype activation. One can observe that instances of a negative bag usually activate negative prototypes (four upper prototypes in red brackets), while the instances of positive bags mostly activate positive prototypes (four bottom prototypes in green brackets).}
  \label{fig:mnist_small_matrix}
\end{figure*}

We experiment on two histological datasets as out toy task: Colon Cancer and Bisque breast cancer. The former contains $100$ H\&E images with $22,444$ manually annotated nuclei of four different types: epithelial, inflammatory, fibroblast, and miscellaneous. To create bags of instances, we extract $27\times27$ nucleus-centered patches from each image, and the goal is to detect if the bag contains one or more epithelial cells, as colon cancer originates from them. On the other hand, the Bisque dataset consists of $58$ H\&E breast histology images of size $896\times768$, out of which $32$ are benign, and $26$ are malignant (contain at least one cancer cell). Each image is divided into $32\times32$ patches, resulting in $672$ patches per image. Patches with at least $75\%$ of the white pixels are discarded, resulting in $58$ bags of various sizes.

We apply extensive data augmentation for both datasets, including random rotations, horizontal and vertical flipping, random staining augmentation, staining normalization, and instance normalization. We use ResNet-18 convolutional parts with the first layer modified to $3\times3$ convolution with stride $1$ to match the size of smaller instances. We set the number of prototypes per class to $10$ with a size of $128\times2\times2$. Warmup, fine-tuning, and end-to-end training take $60$, $20$, and $20$ epochs, respectively. $10$-fold cross-validation with $1$ validation fold and $1$ test fold is repeated $5$ times.

\begin{table*}
\centering
\scriptsize % \scriptsize or \footnotesize or \small
\begin{tabular}{|c||c|c|}
\hline
& \multicolumn{2}{c|}{\textsc{Colon Cancer}} \\ [0.5ex]
\cline{2-3}
\textsc{Method} & \textsc{Accuracy} & \textsc{AUC} \\ [0.5ex]
\hline
\textsc{instance+max*} & $84.2\%\pm 2.1\%$ & $0.914\pm 0.010$ \\ [1ex]
\textsc{instance+mean*} & $77.2\%\pm 1.2\%$ & $0.866\pm 0.008$ \\ [1ex]
\textsc{embedding+max*} & $82.4\%\pm 1.5\%$ & $0.918\pm 0.010$ \\ [1ex]
\textsc{embedding+mean*} & $86.0\%\pm 1.4\%$ & $0.940\pm 0.010$\\ [1ex]
\textsc{AbMILP*} & $88.4\%\pm 1.4\%$ & $0.973\pm 0.007$ \\ [1ex]
\textsc{SA-AbMILP**} & \textbf{90.8\%} $\pm$ \textbf{1.3\%} & \textbf{0.981} $\pm$ \textbf{0.007} \\ [1ex]
\hline
\textsc{ProtoMIL (our)} & $81.3\%\pm 1.9\%$ & $0.932\pm 0.014$\\ [1ex]
\hline
\end{tabular}
\caption{Results for Colon Cancer dataset. ProtoMIL achieves slightly worse results for the Colon Cancer dataset, probably due to its small size. Notice that values for comparison indicated with ``*'' and ``**'' comes from~\cite{ilse2018attention} and~\cite{rymarczyk2021kernel}, respectively.}
\label{table:bisque_colon}
\end{table*}

\paragraph{Results.}
Table \ref{table:bisque_colon} presents our results compared to both traditional and attention-based MIL models. On the Bisque dataset, our model significantly outperforms all baseline models. However, due to the small size of the Colon Cancer dataset, ProtoMIL overfits, resulting in poorer AUC than attention-based models. Nevertheless, in both cases, ProtoMIL provides finer explanations than all baseline models (see Figure~\ref{fig:mnist_small_matrix}b and Supplementary Materials).

\begin{table}[t]
\centering
\footnotesize % \scriptsize or \footnotesize or \small
\begin{tabular}{|c|c|c|}
\hline
\textsc{Method} & \textsc{Accuracy} & \textsc{F-score} \\ [0.5ex]
\hline
\textsc{MI-SVM*} & 54.5\% & 0.70 \\ [1ex]
\textsc{mi-SVM*} & 54.5\% & 0.71 \\ [1ex]
\textsc{EMDD*} & 55.1\% & 0.69 \\ [1ex]
\textsc{Citation k-NN*} & 62.8\% & 0.69 \\ [1ex]
\textsc{MILBoost*} & 64.1\% & 0.66 \\ [1ex]
\textsc{mi-Graph*} & 72.5\% & 0.75 \\ [1ex]
\textsc{MIL-GNN-Att*} & 72.9\% & 0.75 \\ [1ex]
\textsc{MIL-GNN-DP*} & 74.2\% & \textbf{0.77} \\ [1ex]
\textsc{AbMILP**} & 74.5\% & 0.74 \\ [1ex]
\textsc{SA-AbMILP**} & 75.2\% & 0.76 \\ [1ex]
\textsc{LSA-AbMILP**} & \textbf{76.3\%} & \textbf{0.77} \\ [1ex]
\hline
\textsc{ProtoMIL (our)} & 70.0\% & 0.75 \\ [1ex]
\hline
\end{tabular}
\caption{Results for the Messidor dataset show that in terms of F-score, our ProtoMIL method is comparable with methods based on attention (AbMILP) or graph convolutions (MIL-GNN-ATT). Notice that values for comparison marked with ``*'' and ``**'' are taken from~\cite{tu2019multiple} and~\cite{rymarczyk2021kernel}, respectively.}
\label{table:data2}
\end{table}

\begin{figure}[t]
\centering
\includegraphics[width=0.38\textwidth]{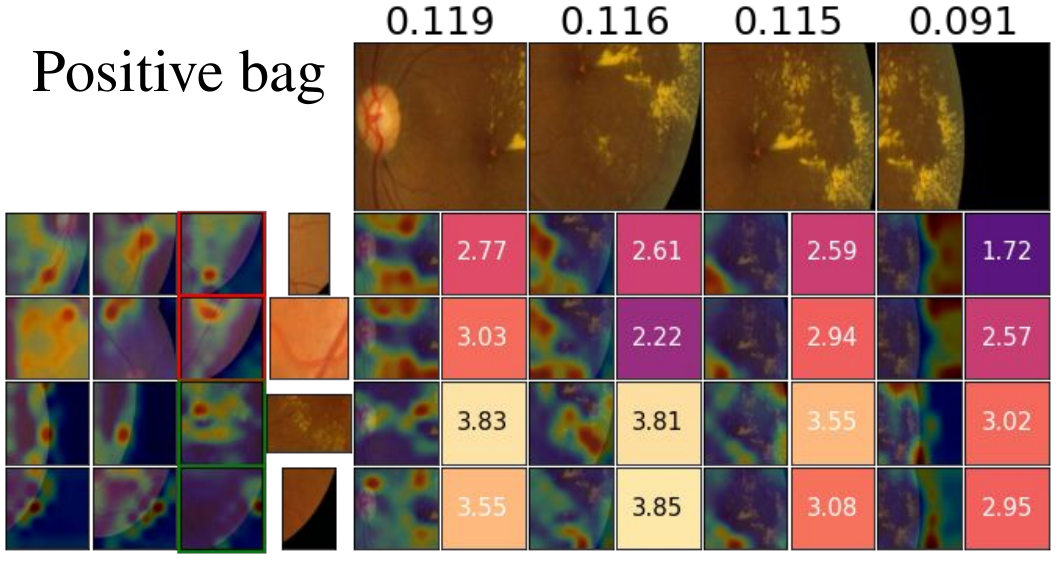}
\caption{Similarity scores between four crucial instances of a bag (columns) and four prototypical parts (rows) for a positive bag from the Messidor dataset. One can observe that ProtoMIL focuses on the disease factors, which are the brightest yellow spots on the image. Moreover, both positive and negative prototypes are activated since the retina with pathological changes still shows healthy features, such as veins. Please refer to Figure~\ref{fig:mnist_small_matrix} for a detailed description of the visualization.}
\label{fig:messidor_short_matrix}
\end{figure}

\subsection{Messidor dataset}
\paragraph{Experiment details.}
The Messidor dataset contains $1200$ retinal images: $654$ with a positive label (diabetic retinopathy) and $546$ with a negative one. To create bags of instances, we crop overlapping patches of size $224\times224$ from each of $700\times700$ images, and patches with more than $70\%$ black pixels are dropped as in~\cite{tu2019multiple}. Additionally, we apply extensive data augmentation, including random rotations, horizontal and vertical flipping, Gaussian noise, and patch normalization. We use ResNet-18 convolutional layers learned from scratch with $10$ prototypes per class and prototype size of $1\times1\times128$. Warmup, fine-tuning, and end-to-end training take $30$, $20$, and $10$ epochs, respectively. We perform 10 fold cross-validation repeated two times as in~\cite{tu2019multiple}.

\paragraph{Results.}
Results of ProtoMIL  in the case of F-score are comparable with the ones achieved in~\cite{tu2019multiple} and~\cite{rymarczyk2021kernel} (see Table~\ref{table:data2}). However, the accuracy is significantly lower, most possibly due to the data class imbalance. Nevertheless, our model provides a fine-grained interpretation of its decision, as presented in Figure~\ref{fig:messidor_short_matrix}. 

\subsection{Additional pruning results}

\begin{table*}[ht!]
\centering
\scriptsize % \scriptsize or \footnotesize or \small
\begin{tabular}{|c||c|c|c||c|c|c|}
\hline
& \multicolumn{3}{c||}{\textsc{Before pruning}} & \multicolumn{3}{c|}{\textsc{After pruning}} \\ [0.5ex]
\cline{2-7}
\textsc{Dataset} & \textsc{Proto. \#} & \textsc{Accuracy} & \textsc{AUC} & \textsc{Proto. \#} & \textsc{Accuracy} & \textsc{AUC} \\ [0.5ex]
\hline
\textsc{MNIST Bags 500} & 20 $\pm$ 0 & 99.2\% $\pm$ 0.1\% & 0.999 $\pm$ 0.001 & 14.12 $\pm$ 0.28 & 99.2\% $\pm$ 0.1\% & 0.999 $\pm$ 0.001 \\ [1ex]
% \textsc{Bisque} & 20 $\pm$ 0 & 76.7\% $\pm$ 2.2\% & 0.886 $\pm$ 0.033 & 13.6 $\pm$ 0.25 & 73.0\% $\pm$ 2.4\% & 0.867 $\pm$ 0.022 \\ [1ex]
% \textsc{Colon Cancer} & 20 $\pm$ 0 & 81.3\% $\pm$ 1.9\% & 0.932 $\pm$ 0.014 & 15.69 $\pm$ 0.34 & 81.8\% $\pm$ 2.4\% & 0.880 $\pm$ 0.022\\ [1ex]
\textsc{Messidor} & 20 $\pm$ 0 & 70.0\% $\pm$ 0.9\% & 0.692 $\pm$ 0.012 & 16.70 $\pm$ 1.86 & 64.7\% $\pm$ 1.3\% & 0.717 $\pm$ 0.013 \\ [1ex]
% \textsc{Camelyon16} & 10 $\pm$ 0 & 87.3\% $\pm$ 1.2 \% & 0.935 $\pm$ 0.007 & 6.4 $\pm$ 0.24 & 85.9\% $\pm$ 1.5\% & 0.937 $\pm$ 0.007 \\ [1ex]
% \textsc{TCGA-NSCLC} & 10 $\pm$ 0 & -\% $\pm$  -\% & 0.918 $\pm$ 0.003 & 7.6 $\pm$ 1.2 & 81.1\% $\pm$ 1.4\% & 0.880 $\pm$ 0.003  \\ [1ex]
% \textsc{TCGA-RCC} & 10 $\pm$ 0 & 94.66\% $\pm$  1.0\% & 0.988 $\pm$ 0.009 & 6.5 $\pm$ 1.5 & 91.0\% $\pm$ 1.6\% & 0.955 $\pm$ 0.006  \\ [1ex]
\hline
\end{tabular}
\caption{The influence of ProtoMIL pruning on the accuracy and AUC score. One can notice that even though the pruning removes around $30\%$ of the prototypes, it usually does not noticeably decrease the AUC and accuracy of the model.}

\label{table:pruning}
\end{table*}

\section{Additional visualizations}
\begin{figure*}[ht]
  \centering
  \includegraphics[width=\textwidth,trim={10cm 8cm 8cm 8cm},clip]{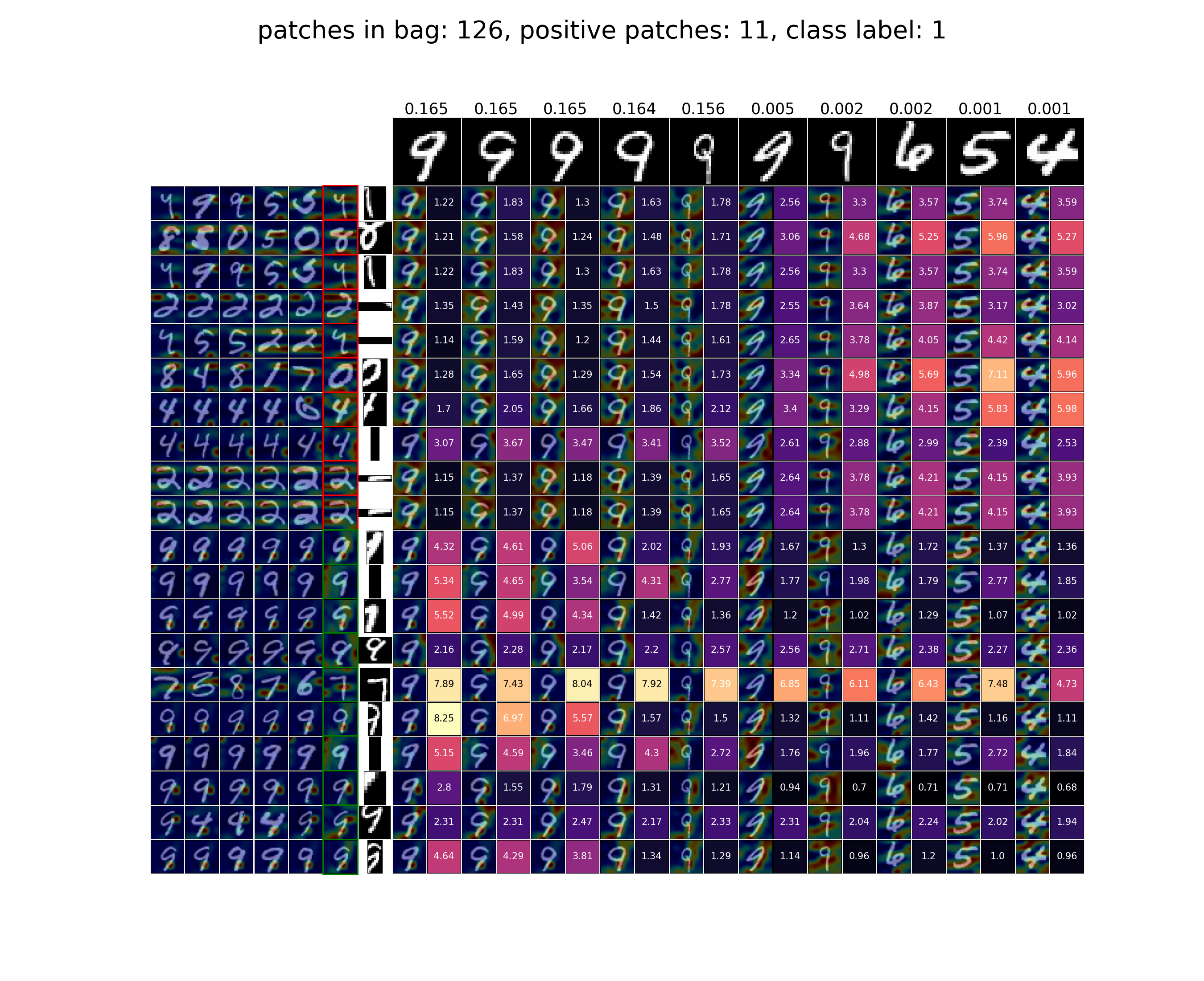}
  \caption{Similarity scores for a positive bag from MNIST Bags.}
  \label{fig:mnist_positive_matrix}
\end{figure*}

\begin{figure*}[ht]
  \centering
  \includegraphics[width=\textwidth,trim={10cm 8cm 8cm 8cm},clip]{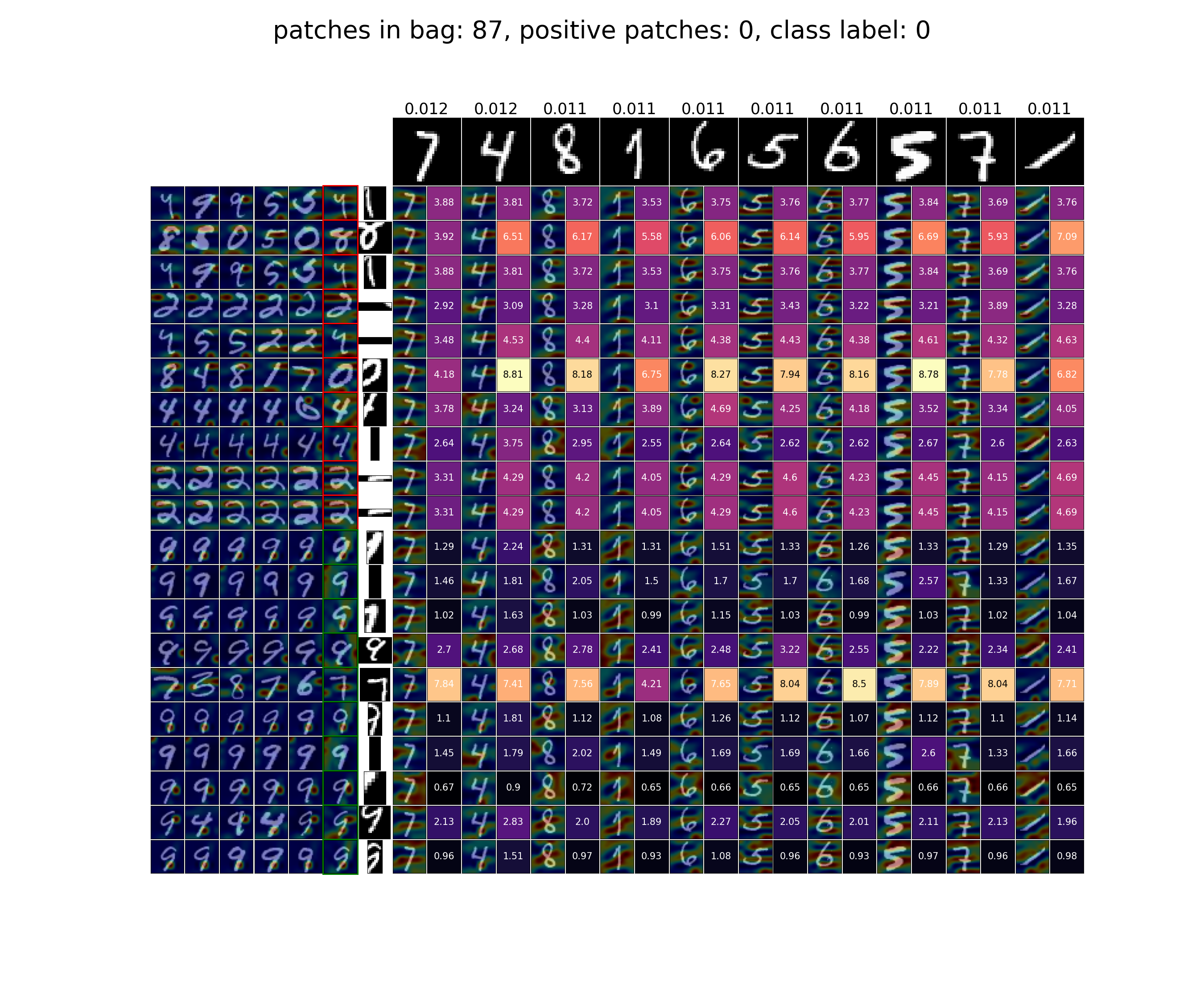}
  \caption{Similarity scores for a negative bag from MNIST Bags.}
  \label{fig:mnist_negative_matrix}
\end{figure*}

\begin{figure*}[ht]
  \centering
  \includegraphics[width=\textwidth,trim={10cm 8cm 8cm 8cm},clip]{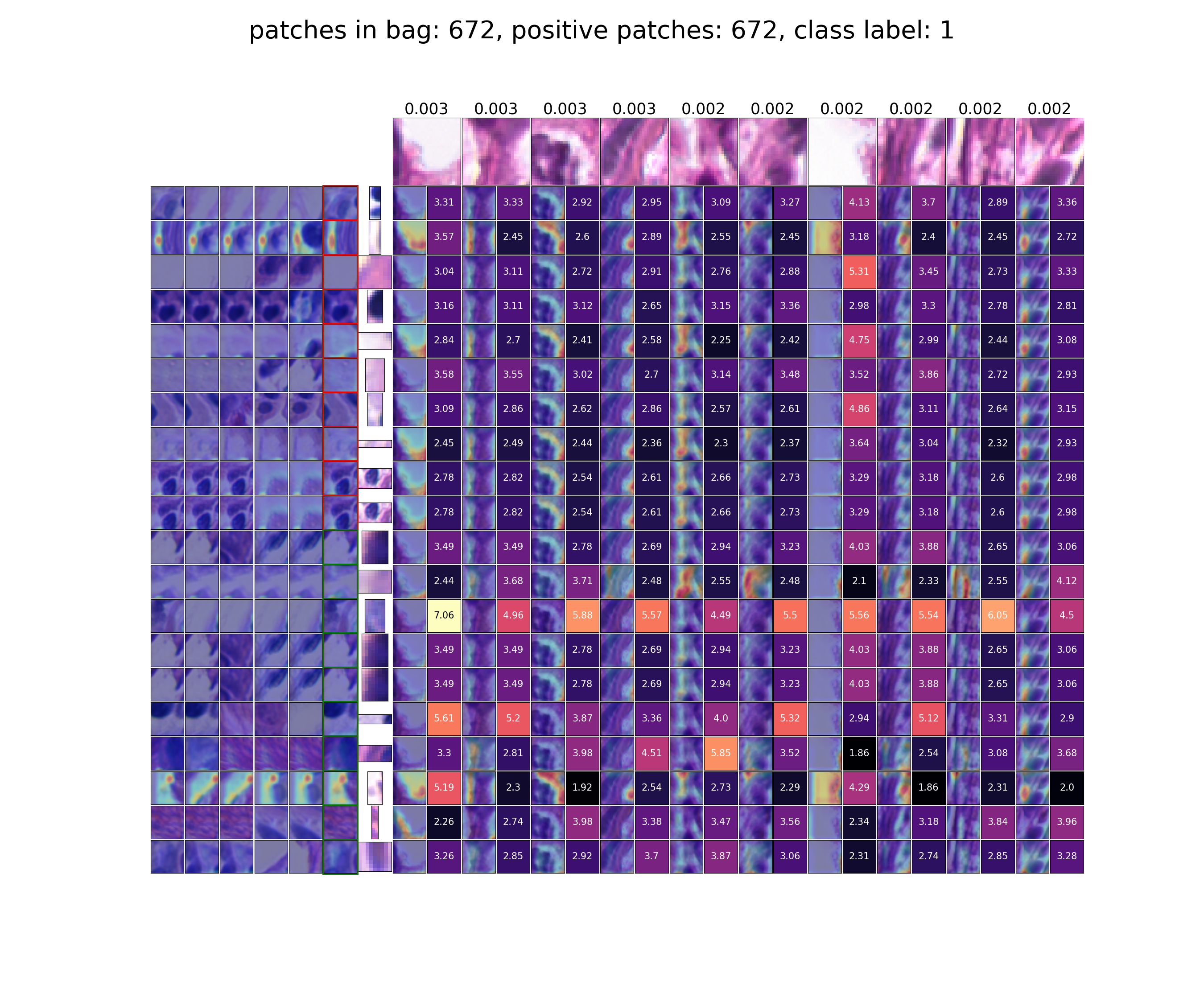}
  \caption{Similarity scores for a positive bag from Bisque dataset.}
  \label{fig:bisque_positive_matrix}
\end{figure*}

\begin{figure*}[ht]
  \centering
  \includegraphics[width=\textwidth,trim={10cm 8cm 8cm 8cm},clip]{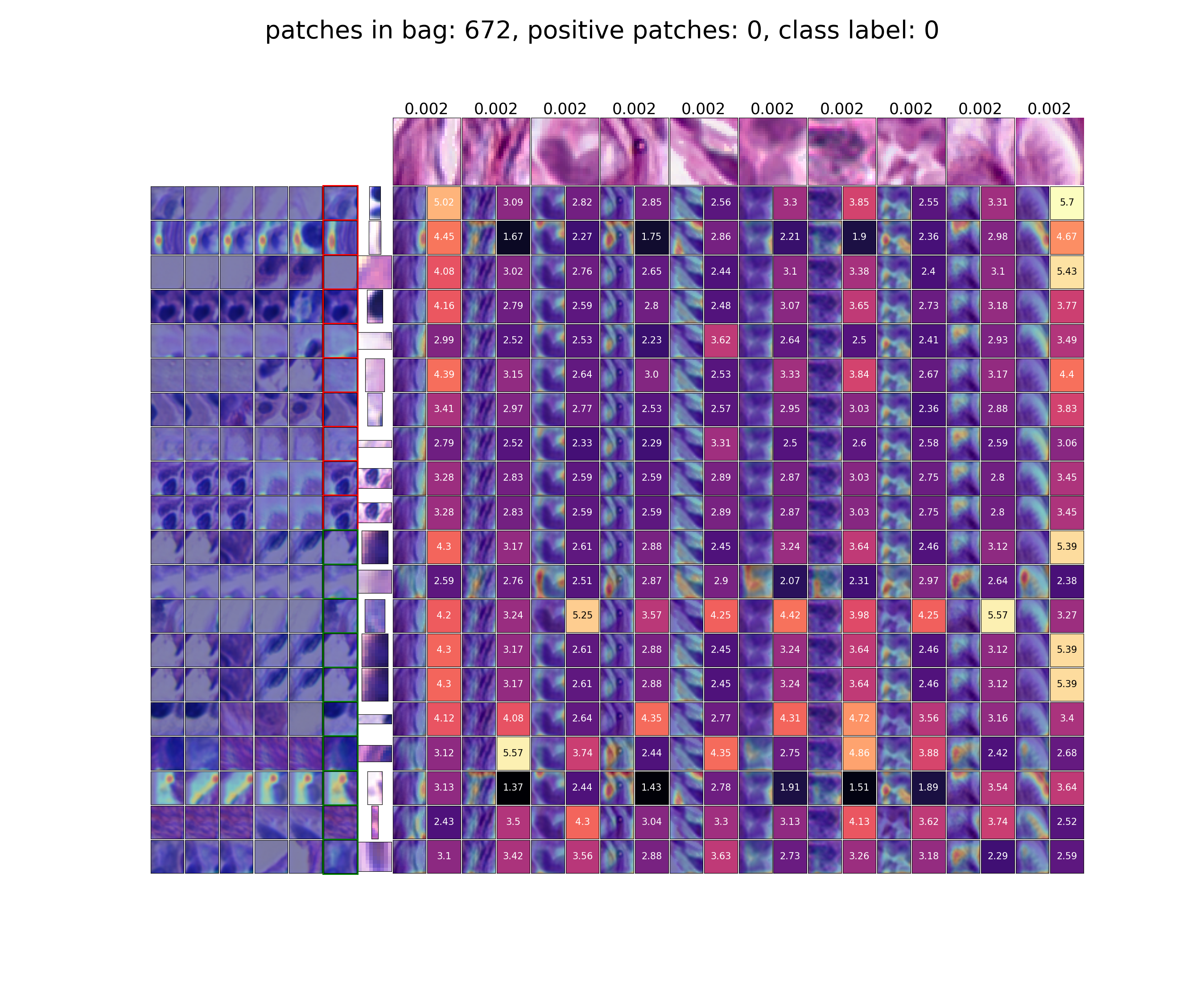}
  \caption{Similarity scores for a negative bag from Bisque dataset.}
  \label{fig:bisque_negative_matrix}
\end{figure*}

\begin{figure*}[ht]
  \centering
  \includegraphics[width=\textwidth,trim={10cm 8cm 8cm 8cm},clip]{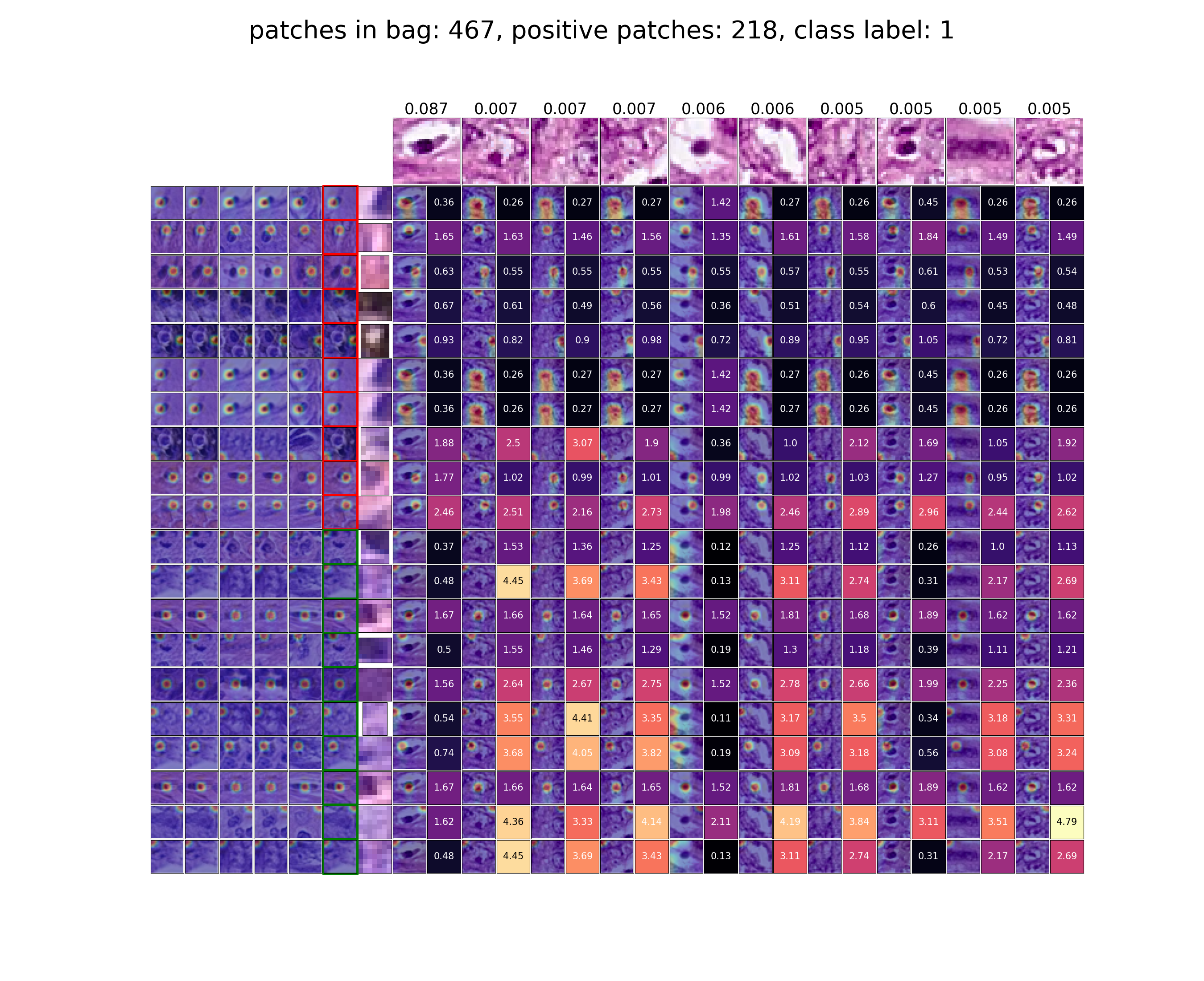}
  \caption{ProtoMIL analysis matrix for a positive example from Colon Cancer dataset.}
  \label{fig:colon_positive_matrix}
\end{figure*}

\begin{figure*}[ht]
  \centering
  \includegraphics[width=\textwidth,trim={10cm 8cm 8cm 8cm},clip]{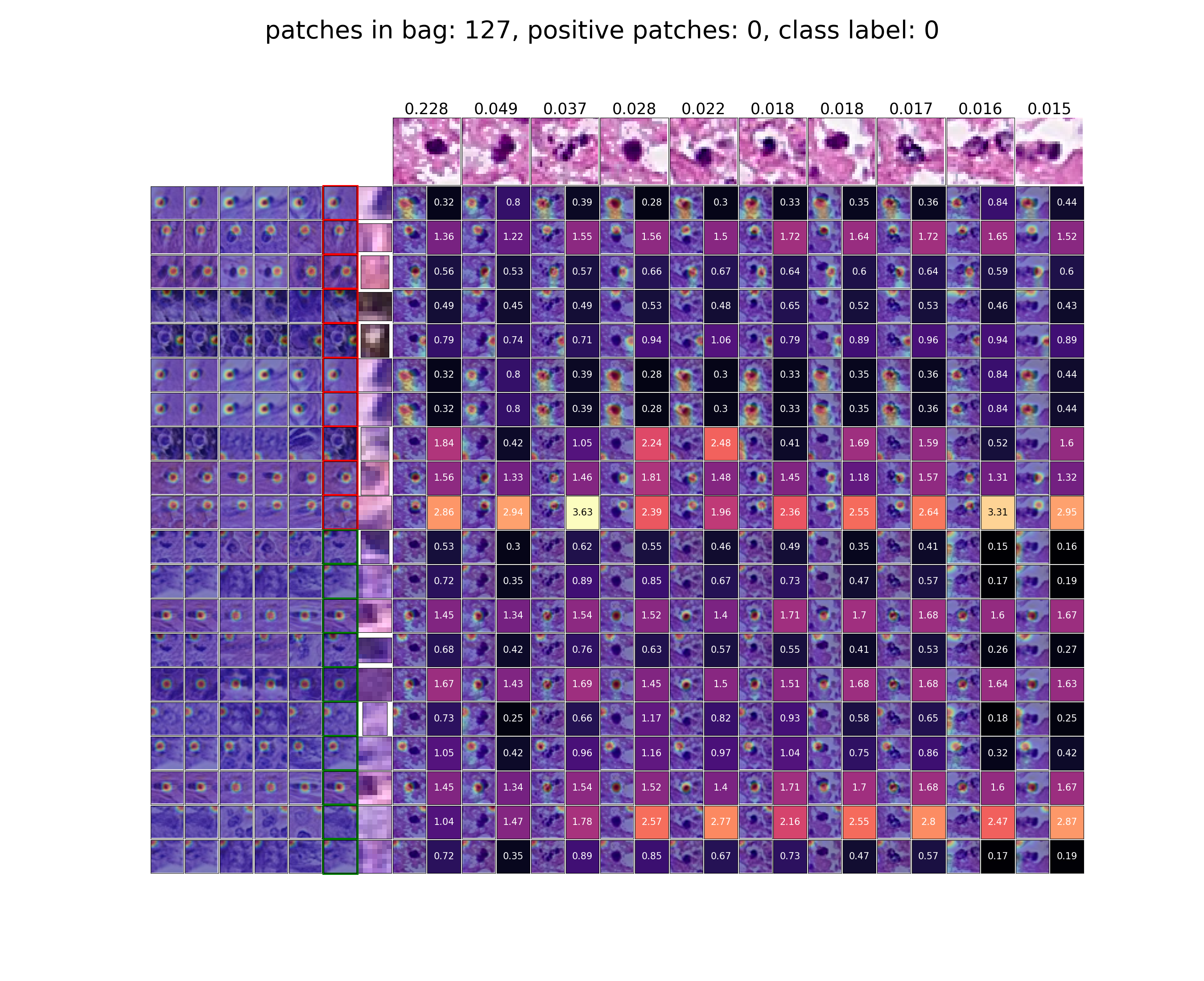}
  \caption{ProtoMIL analysis matrix for a negative example from Colon Cancer dataset.}
  \label{fig:colon_negative_matrix}
\end{figure*}

\begin{figure*}[ht]
  \centering
  \includegraphics[width=\textwidth]{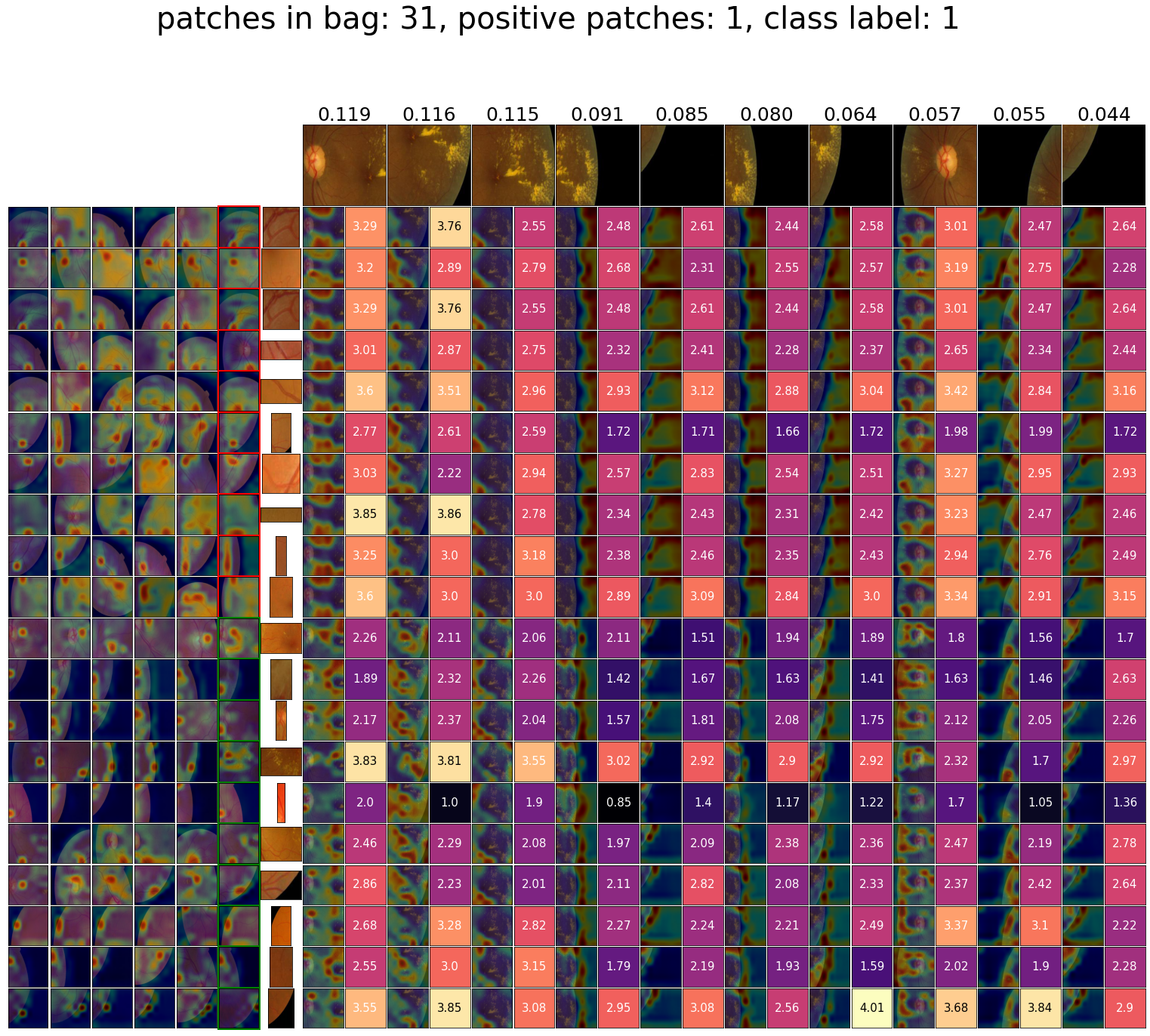}
  \caption{Similarity scores for a positive bag from Messidor dataset.}
  \label{fig:camelyon_positive_matrix}
\end{figure*}

\begin{figure*}[ht]
  \centering
  \includegraphics[width=\textwidth,trim={6cm 3cm 4cm 2cm},clip]{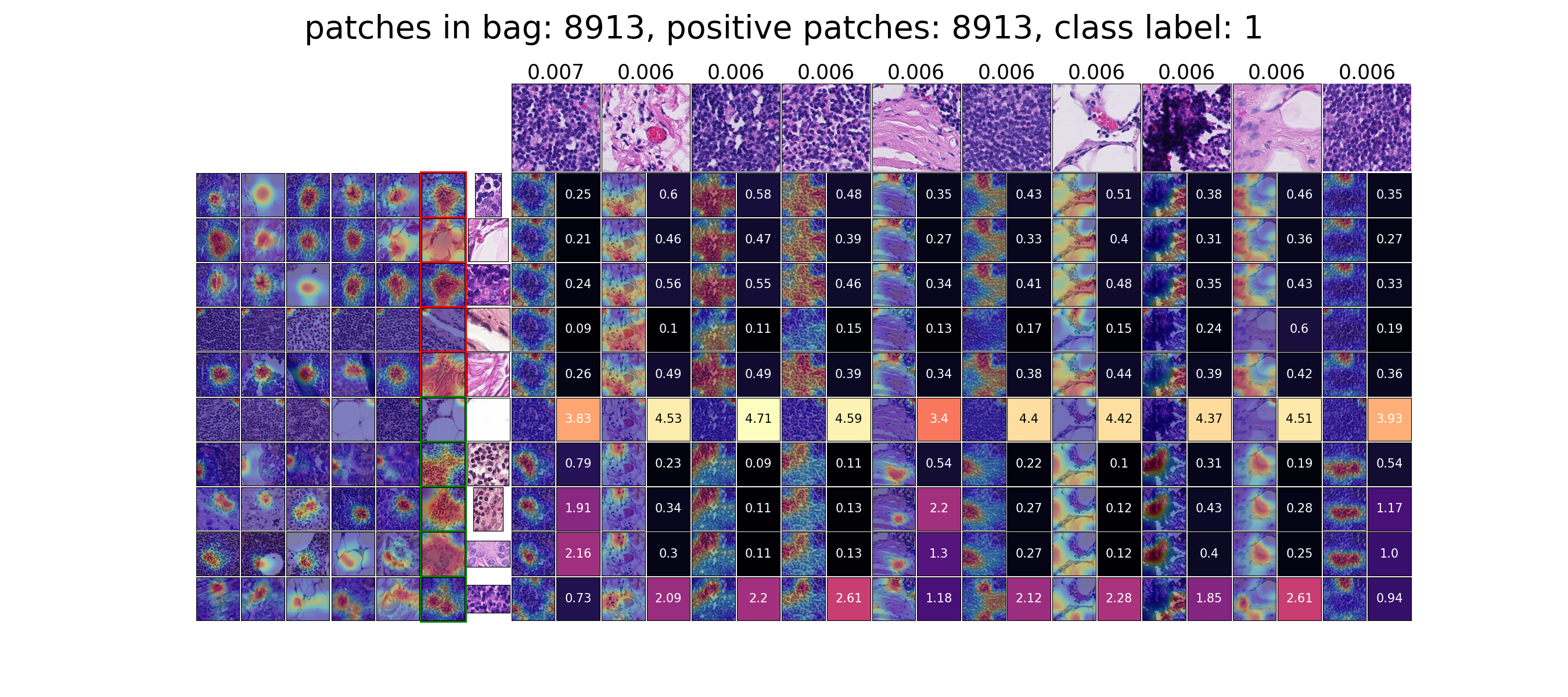}
  \caption{Similarity scores for a positive bag from Camelyon16 dataset.}
  \label{fig:camelyon_positive_matrix}
\end{figure*}

\begin{figure*}[ht]
  \centering
  \includegraphics[width=\textwidth,trim={6cm 3cm 4cm 2cm},clip]{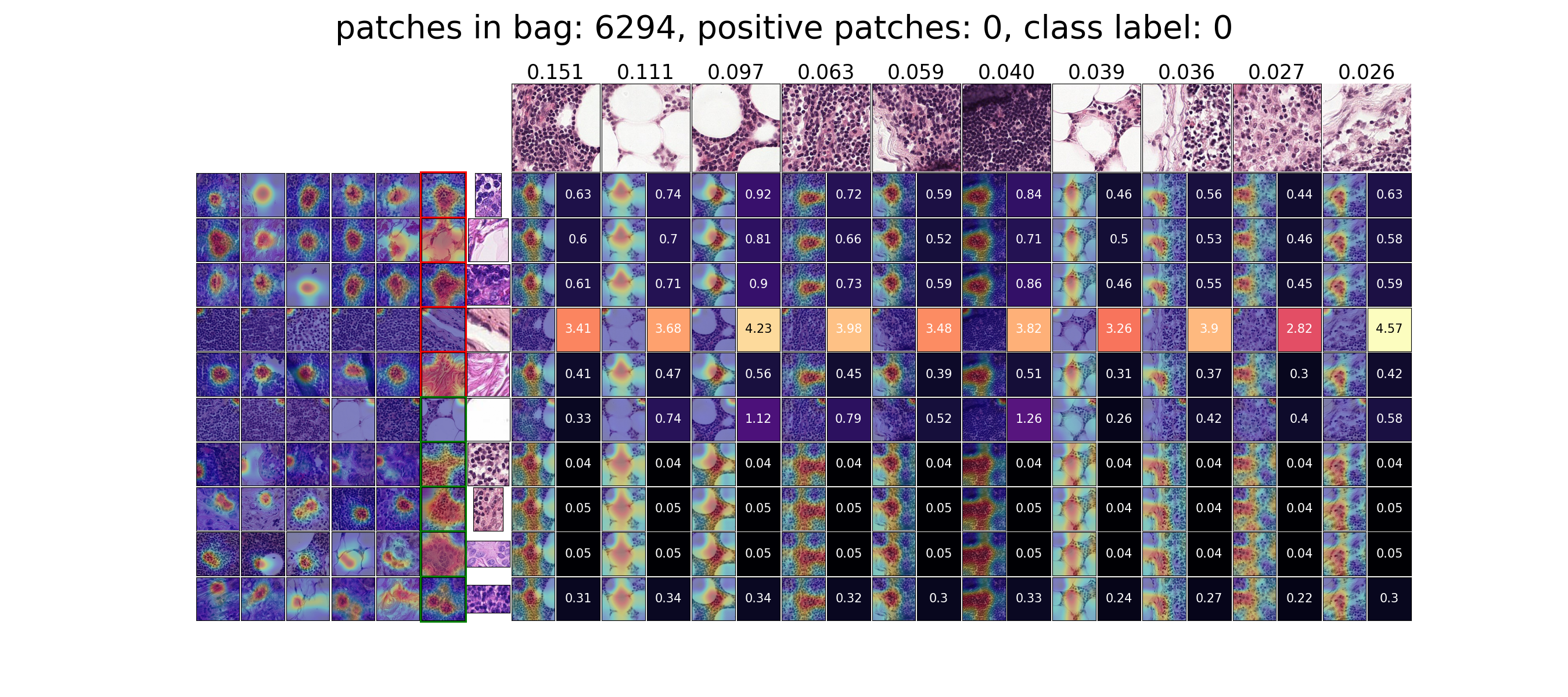}
  \caption{Similarity scores for a negative bag from Camelyon16 dataset.}
  \label{fig:camelyon_negative_matrix}
\end{figure*}

\begin{figure*}[ht]
  \centering
  \includegraphics[width=0.91\textwidth]{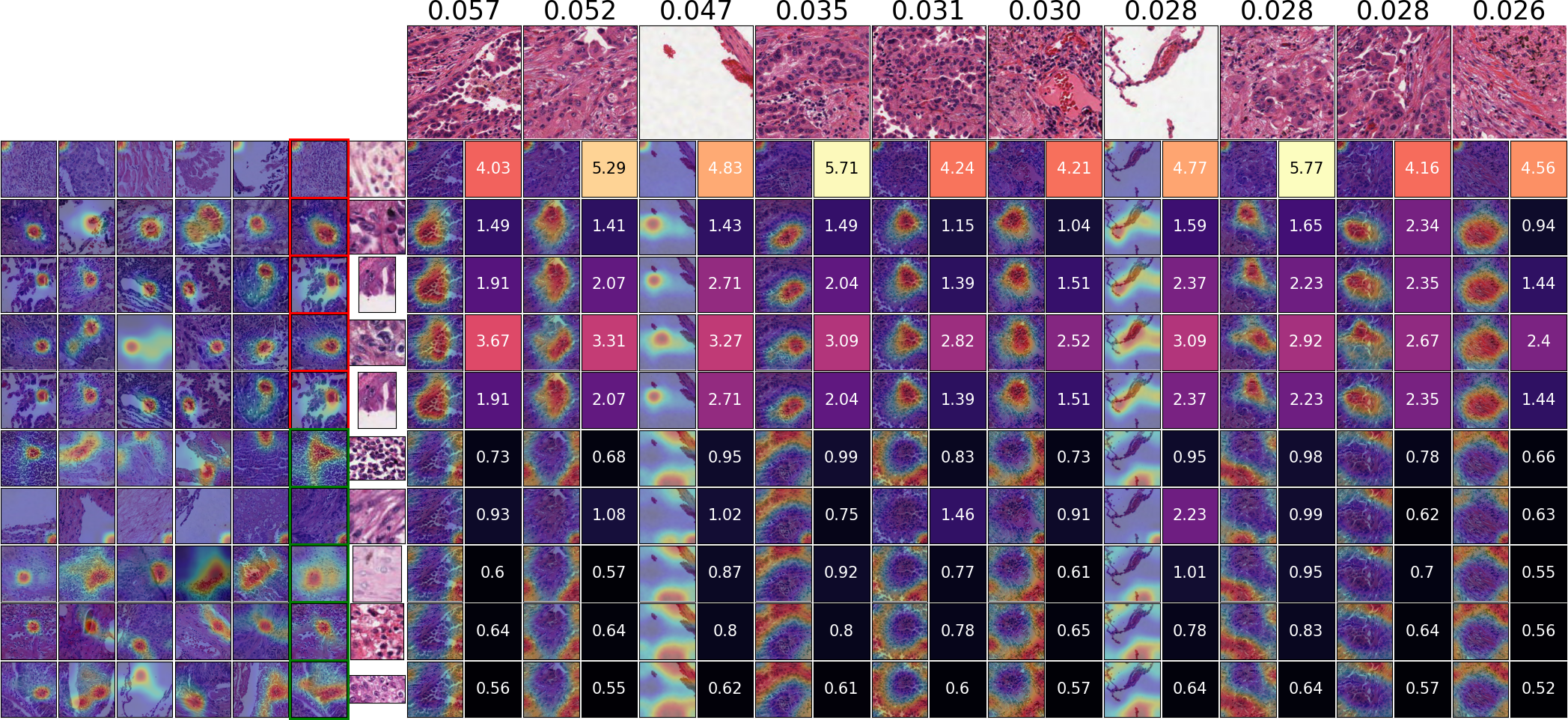}
  \caption{Similarity scores for a LUAD bag from TCGA-NSCLC dataset.}
  \label{fig:nsclc_luad_matrix}
\end{figure*}

\begin{figure*}[ht]
  \centering
  \includegraphics[width=0.91\textwidth]{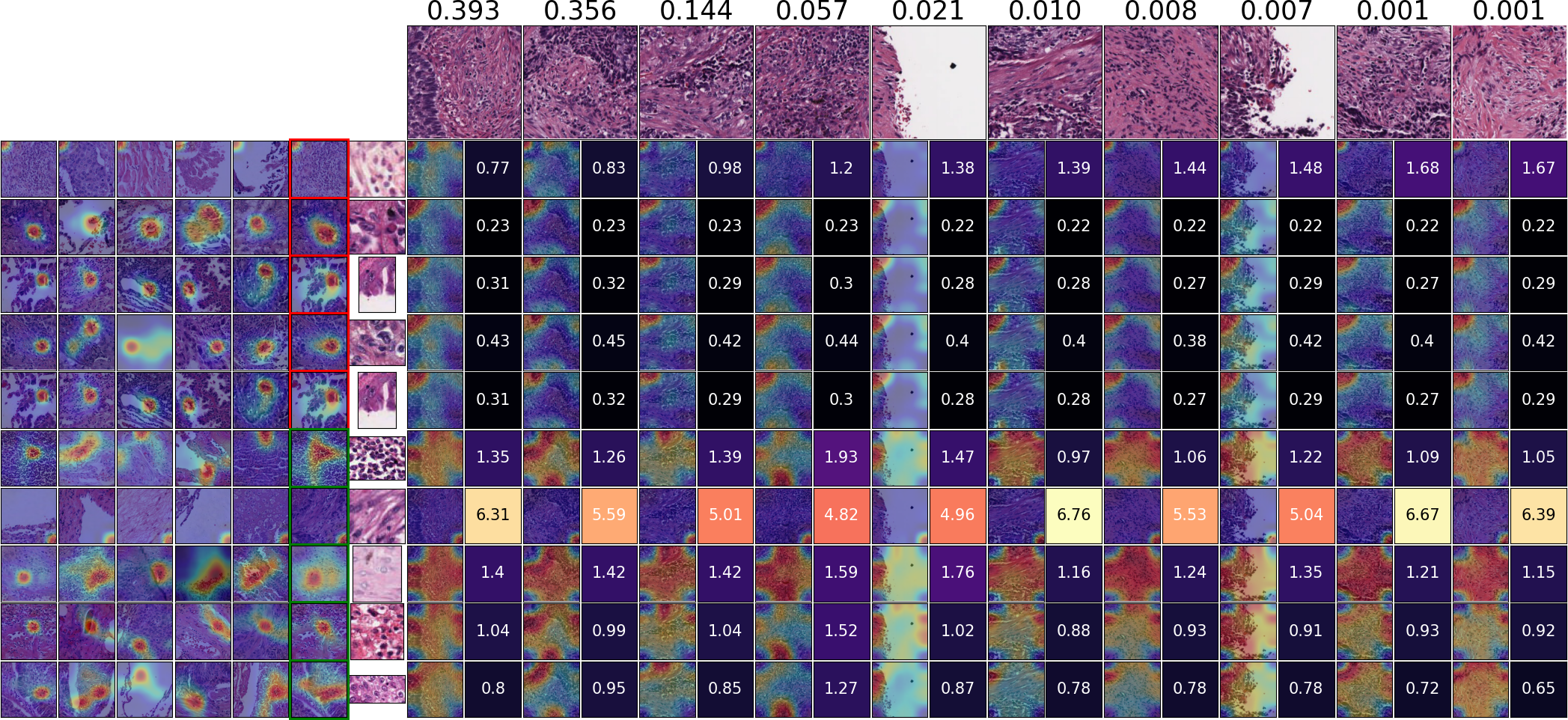}
  \caption{Similarity scores for a LUSC bag from TCGA-NSCLC dataset.}
  \label{fig:camelyon_negative_matrix}
\end{figure*}

\begin{figure*}[ht]
  \centering
  \includegraphics[width=\textwidth,trim={8cm 3cm 5cm 3cm},clip]{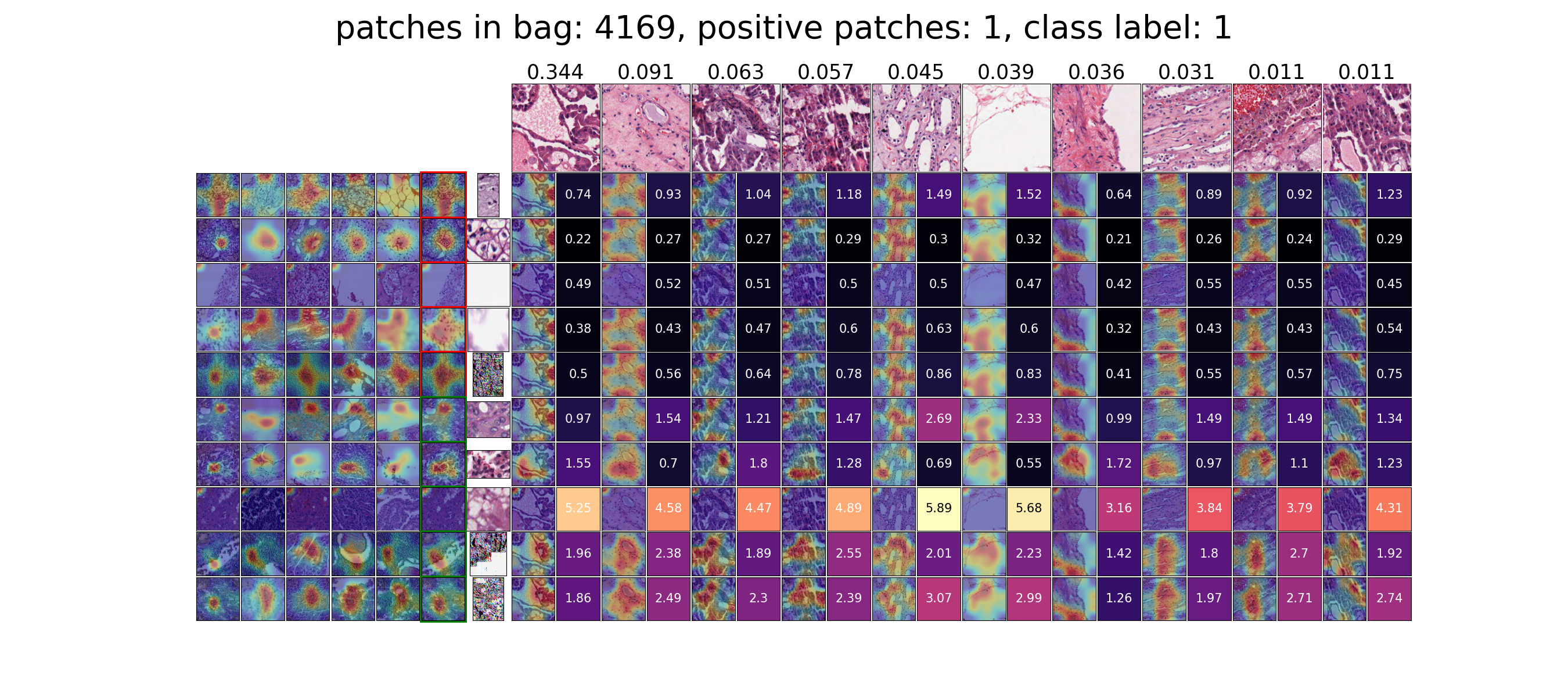}
  \caption{Similarity scores for a positive bag from TCGA RCC dataset.}
  \label{fig:camelyon_positive_matrix}
\end{figure*}

\begin{figure*}[ht]
  \centering
  \includegraphics[width=\textwidth,trim={8cm 3cm 5cm 3cm},clip]{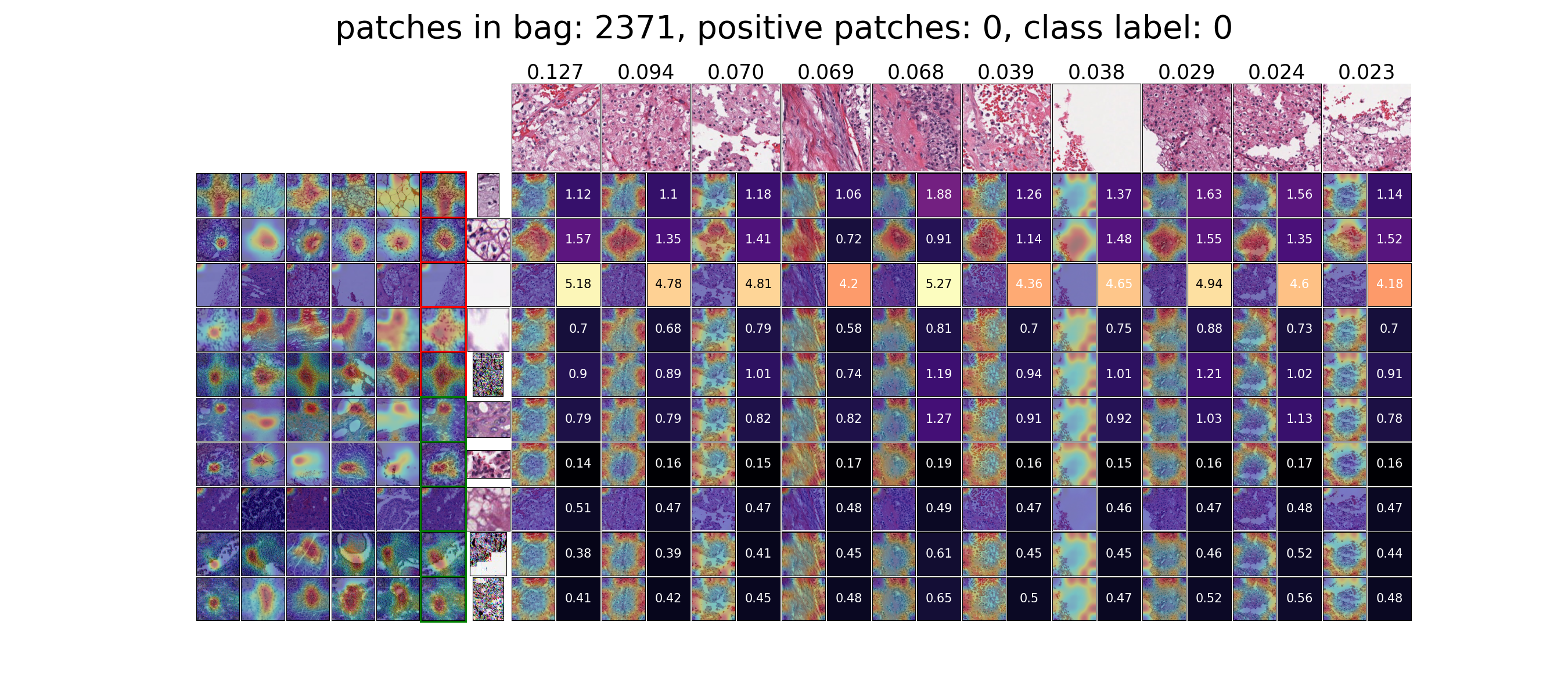}
  \caption{Similarity scores for a negative bag from TCGA RCC dataset.}
  \label{fig:camelyon_negative_matrix}
\end{figure*}

\end{document}